\documentclass{article} 
\usepackage{iclr2022_conference,times}

\DeclareMathAlphabet{\mathsfit}{\encodingdefault}{\sfdefault}{m}{sl}
\SetMathAlphabet{\mathsfit}{bold}{\encodingdefault}{\sfdefault}{bx}{n}

\def\sR{{\mathbb{R}}}

\usepackage{amsfonts}
\usepackage{amsmath,bm}

\newtheorem{theorem}{Theorem}
\newtheorem{lemma}{Lemma}
\newtheorem{corollary}{Corollary}

\newtheorem{proposition}{Proposition}
\newtheorem{assum}{Assumption}

\usepackage{xargs}                      
\usepackage[colorinlistoftodos,prependcaption]{todonotes}
\newcommandx{\red}[2][1=]{\todo[linecolor=red,backgroundcolor=red!25,bordercolor=red,#1]{#2}}
\newcommandx{\blue}[2][1=]{\todo[linecolor=blue,backgroundcolor=blue!25,bordercolor=blue,#1]{#2}}
\newcommandx{\green}[2][1=]{\todo[linecolor=OliveGreen,backgroundcolor=OliveGreen!25,bordercolor=OliveGreen,#1]{#2}}
\newcommandx{\purple}[2][1=]{\todo[linecolor=Plum,backgroundcolor=Plum!25,bordercolor=Plum,#1]{#2}}

\usepackage{mathrsfs}

\newcommand{\del}{\mathrm{d}}

\newcommand{\lp}{\left(}
\newcommand{\rp}{\right)}

\usepackage{graphicx,wrapfig}
\usepackage{hyperref}
\usepackage{url}
\usepackage[ruled,vlined]{algorithm2e}
\usepackage{amssymb}
\usepackage{comment}

\title{Graphon based Clustering and Testing of Networks: Algorithms and Theory}

\author{Mahalakshmi Sabanayagam\\
Technical University of Munich\\
\And
Leena Chennuru Vankadara \\
IMPRS-IS, University of T{\"u}bingen
\AND
Debarghya Ghoshdastidar \\
Technical University of Munich

}

\iclrfinalcopy 
\begin{document}

\maketitle

\begin{abstract}
Network-valued data are encountered in a wide range of applications, and pose challenges in learning due to their complex structure and absence of vertex correspondence. Typical examples of such problems include classification or grouping of protein structures and social networks.
Various methods, ranging from graph kernels to graph neural networks, have been proposed that achieve some success in graph classification problems.
However, most methods have limited theoretical justification,
and their applicability beyond classification remains unexplored.
In this work, we propose methods for clustering multiple graphs, without vertex correspondence, that are inspired by the recent literature on estimating graphons---symmetric functions corresponding to infinite vertex limit of graphs. 
We propose a novel graph distance based on sorting-and-smoothing graphon estimators.
Using the proposed graph distance, we present two clustering algorithms and show that they achieve state-of-the-art results. 
We prove the statistical consistency of both algorithms under Lipschitz assumptions on the graph degrees.
We further study the applicability of the proposed distance for graph two-sample testing problems.

\end{abstract}

\section{Introduction}

Machine learning on graphs has evolved considerably over the past two decades.
The traditional view towards network analysis is limited to modelling interactions among entities of interest, for instance social networks or world wide web, and learning algorithms based on graph theory have been commonly used to solve these problems \citep{von2007tutorial,yan2006graph}.
However, recent applications in bioinformatics and other disciplines require a different perspective, where the networks are the quantities of interest.
For instance, it is of practical interest to classify protein structures as enzyme or non-enzyme~\citep{dobson2003distinguishing} or detect topological changes in brain networks caused by Alzheimer's disease \citep{stam2007small}. 
We refer to such problems as learning from network-valued data to distinguish from the traditional network analysis problems, involving a single network of interactions \citep{newman2003structure}.

Machine learning on network-valued data has been an active area of research in recent years, although most works focus on the network classification problem.
The generic approach is to convert the network-valued data into a standard representation.
Graph neural networks are commonly used for network embedding, that is, finding Euclidean representations of each network that can be further used in standard machine learning models   \citep{narayanan2017graph2vec,xu2019powerful}.
In contrast, graph kernels capture similarities between pairs of networks that can be used in kernel based learning algorithms \citep{shervashidze2011weisfeiler,kondor2016multiscale,togninalli2019wasserstein}. 
In particular, the graph neural tangent kernel defines a graph kernel that corresponds to infinitely wide graph neural networks, and typically outperforms neural networks in classification tasks \citep{du2019graph}.
A more classical equivalent for graph kernels is to define metrics that characterise the distances between pairs of graphs \citep{bunke1998graph}, but there has been limited research on designing efficient and useful graph distances in the machine learning literature.

The motivation for this paper stems from two shortcomings in the literature on network-valued data analysis: first, the efficacy of existing kernels or embeddings have not been studied beyond network classification, and second is the lack of theoretical analysis of these methods, particularly in the small sample setting.
Generalisation error bounds for graph kernel based learning exist \citep{du2019graph}, but these bounds, based on learning theory, are meaningful only when many networks are available. However, in many applications, one needs to learn from a small population of large networks and, in such cases, an informative statistical analysis should consider the small sample, large graph regime.
To address this issue, we take inspiration from the recent statistics literature on graph two-sample testing---given two (populations of) large graphs, the goal is to decide if they are from same statistical model or not.
Although most theoretical studies in graph two-sample testing focus on graph with vertex correspondence \citep{tang2017semiparametric,ghoshdastidar2018practical}, some works address the problem of testing graphs on different vertex sets either by defining distances between graphs \citep{tang2017nonparametric,agterberg2020nonparametric} or by representing networks in terms of pre-specified network statistics \citep{ghoshdastidar2017two}. 
The use of network statistics for clustering network-valued data is studied in \citet{mukherjee2017clustering}.
Another fundamental approach for dealing with graphs of different sizes is graph matching, where the objective is to determine the vertex correspondence.
Graph matching is often solved by formulating it as an optimization problem \citep{zaslavskiy2008path, guo2019quotient} or defining graph edit distance between the graphs \citep{riesen2009approximate,gao2010survey}.
Although, there is extensive research on graph matching, the efficacy of these methods in learning from network-valued data remains unexplored.

\textbf{Contribution and organisation.} 
In this work, we follow the approach of defining meaningful graph distances based on statistical models, and use the proposed graph distance in the context of learning from networks without vertex correspondence. 
In particular, we propose graph distances based on graphons. Graphons are symmetric bivariate functions that represent the limiting structure for a sequence of graphs with increasing number of nodes \citep{lovasz2006limits}, but can be also viewed as a nonparametric statistical model for exchangeable random graphs \citep{diaconis2007graph,bickel2009nonparametric}.
The latter perspective is useful for the purpose of machine learning since it allows us to view the multiple graphs as random samples drawn from one or more graphon models. This perspective forms the basis of our contributions, which are listed below:

\textbf{1)}
    In Section~\ref{sec:related_work}, we propose a distance between two networks, that do not have vertex correspondence and could have different number of vertices. We view the networks as random samples from (unknown) graphons, and propose a graph distance that estimates the $L_2$-distance between the graphons. The distance is inspired by the sorting-and-smoothing graphon estimator \citep{chan2014consistent}.

\textbf{2)}
    In Section~\ref{sec:clustering}, we present two algorithms for clustering network-valued data based on the proposed graph distance: a distance-based spectral clustering algorithm, and a similarity based semi-definite programming (SDP) approach.
    We derive performance guarantees for both algorithms under the assumption that the networks are sampled from graphons satisfying certain smoothness conditions.

\textbf{3)}
    We empirically compare the performance of our algorithms with other clustering strategies based on graph kernels, graph matching, network statistics etc. 
    and show that, on both simulated and real data,
    our graph distance-based spectral clustering algorithm outperforms others while the SDP approach also shows reasonable performance, and they also scale to large networks (Section~\ref{sec:experiment}).

\textbf{4)}
    Inspired by the success of the proposed graph distance in clustering, we use the distance for graph two-sample testing. In Section~\ref{sec:hyp_test}, we show that the proposed two-sample test is statistically consistent for large graphs, and also demonstrate the efficacy of the test through numerical simulation.

We provide further discussion in Section~\ref{sec:conclusion} and present the proofs of theoretical results in Appendix.

\section{Graph Distance based on Graphons}
\label{sec:related_work}

Clustering or testing of multiple networks requires a notion of distance between the networks.
In this section, we present a transformation that converts graphs of different sizes into a fixed size representation, and subsequently, propose a graph distance inspired by the theory of graphons. 
We first provide some background on graphons and graphon estimation. 
%
Graphon has been studied in the literature from two perspectives: as limiting structure for infinite sequence of growing graphs \citep{lovasz2006limits}, or as exchangeable random graph model.
In this paper, we follow the latter perspective. A random graph is said to be exchangeable if its distribution is invariant under permutation of nodes.
\citet{diaconis2007graph} showed that any statistical model that generates exchangeable random graphs can be characterised by graphons, as introduced by \citet{lovasz2006limits}.
Formally, a graphon is a symmetric measurable continuous function $w: [0,1]^2 \rightarrow [0,1]$ where $w(x,y)$ can be interpreted as the link probability between two nodes of the graph that are assigned values $x$ and $y$, respectively. 
This interpretation propounds the following two stage sampling procedure for graphons. 
To sample a random graph $G$ with $n$ nodes from a graphon $w$, in the first stage, one samples $n$ variables $U_1,\ldots,U_n$ uniformly from $[0,1]$ and constructs a latent mapping between the sampled points and the node labels. In the second stage, edges between any two nodes $i,j$ are randomly added based on the link probability $w(U_i,U_j)$. 
Mathematically, if we abuse notation to denote the adjacency matrix by $G\in \{0,1\}^{n\times n}$, we have 
\begin{align}
    U_1,\ldots,U_n &\stackrel{\text{iid}}{\sim} \mathit{Uniform}[0,1] 
    \qquad \text{and} \qquad
    G_{ij} | U_i, U_j &\sim \mathit{Bernoulli}(w(U_i, U_j)) \text{ for all } i<j. \nonumber 
\end{align}
We consider problems involving multiple networks sampled independently from the same (or different) graphons. 
We make the following smoothness assumptions on the graphons. 
%
\begin{assum}[Lipschitz continuous] 
\label{assum:lipschitz}
A graphon $w$ is Lipschitz continuous with constant $L$ if
\[ |w(u,v) - w(u',v')| \leq L \sqrt{(u-u')^2+(v-v')^2} \qquad \text{for every } u,v,u',v' \in [0,1].\]
\end{assum}
\begin{assum}[Two-sided Lipschitz degree] 
\label{assum:lipschitz-deg}
A graphon $w$ has two-sided Lipschitz degree with constants $\lambda_1, \lambda_2 > 0$ if its degree distribution $g$, defined by $g(u) = \int_0^1 w(u,v)\mathrm{d}v$, satisfies
$$\lambda_2 \vert u-u' \vert \leq \vert g(u) - g(u') \vert  \leq \lambda_1 \vert u-u' \vert 
\qquad \text{for every } u,u'\in[0,1].$$
\end{assum}
One of the challenges in graphon estimation is due to the issue of non-identifiability, that is, different graphon functions $w$ can generate the same random graph model. In particular, two graphons $w$ and $w'$ generate the same random graph model if they are weakly isomorphic---there exist two measure preserving transformations $\phi, \phi': [0,1] \to [0,1]$ such that $w(\phi(u), \phi(v)) = w'(\phi'(u), \phi'(v))$. Moreover, the converse also holds meaning that such transformations are known to be the only source of non-identifiability \citep{diaconis2007graph}. This weak isomorphism induces equivalence classes on the space of graphons. 
Since our goal is only to cluster graphs belonging to random graph models, we simply make the following assumption on our graphons.
 
 \begin{assum}[Equivalence classes]
\label{assum:equivalence}
Any reference to $K$ graphons, $w_1,\ldots,w_K$, assumes that, for every $i,j$, either $w_i=w_j$ or $w_i$ and $w_j$ belong to different equivalence classes. Furthermore, without loss of generality, we assume that every graphon $w_i$ is represented such that the corresponding degree function $g_i$ is non-decreasing.
\end{assum}

\textbf{Remark on the necessity of Assumptions \ref{assum:lipschitz}--\ref{assum:equivalence}.}
Assumption~\ref{assum:lipschitz} is standard in graphon estimation literature \citep{klopp2017oracle} since it avoids graphons corresponding to inhomogeneous random graph models.
It is known that two graphs from widely separated inhomogeneous models (in $L_2$-distance) are statistically indistinguishable \citep{ghoshdastidar2020two}, and hence, it is essential to ignore such models to derive meaningful guarantees. 
Assumption~\ref{assum:lipschitz-deg} ensures that, under a measure-preserving transformation, the graphon has strictly increasing degree function, which is a canonical representation of an equivalence class of graphons \citep{bickel2009nonparametric}. 
Assumption~\ref{assum:equivalence} is needed since graphons can only be estimated up to measure-preserving transformation. As noted above, it is inconsequential for all practical purposes but simplifies the theoretical exposition. 

\textbf{Graph transformation.}
In order to deal with multiple graphs and measure  distances among pairs of graphs, we require a transformation that maps all graphs into a common metric space---the space of all $n_0\times n_0$ symmetric matrices for some integer $n_0$.
While the graphon estimation literature provides several consistent estimators \citep{klopp2017oracle,zhang2017estimating}, only the histogram based sorting-and-smoothing graphon estimator of \citet{chan2014consistent} can be adapted to meet the above requirement. 
We use the following graph transformation, inspired by \citet{chan2014consistent}. The adjacency matrix $G$ of size $n \times n$ is first reordered based on permutation $\sigma$, such that the empirical degree based on this permutation is monotonically increasing. The degree sorted adjacency matrix is denoted by $G^{\sigma}$. 
It is then transformed to a `histogram' $A\in\sR^{n_0 \times n_0}$ given by
\begin{align}
    A_{ij} = \frac{1}{h^2}\sum_{i_1=1}^{h} \sum_{j_1=1}^{h} G^{\sigma}_{ih+i_1, jh+j_1}, \text{ where $h = \left\lfloor \frac{n}{n_0} \right\rfloor$ and $\lfloor \cdot \rfloor$ is the floor function.}  \label{eq:graph_apprx}
\end{align}

\textbf{Proposed graph distance.}
Given two graphs $G_1$ and $G_2$ with $n_1$ and $n_2$ nodes, respectively, we apply the transformation~\eqref{eq:graph_apprx} to both the  graphs with $n_0 \leq \min\{n_1,n_2\}$. We propose to use the graph distance
\begin{equation}
d(G_1,G_2) = \frac{1}{n_0} \Vert A_1 - A_2\Vert_F,
\label{eq:graphdist}
\end{equation}
where $A_1$ and $A_2$ denote the transformed matrices and $\Vert\cdot\Vert_F$ denotes the matrix Frobenius norm.
Proposition~\ref{prop:grah_dt_conc} shows that, if $G_1$ and $G_2$ are sampled from two graphons, then the graph distance \eqref{eq:graphdist} consistently estimates the $L_2$-distance between the two graphons, which is defined as 
\begin{equation}
   \left \Vert w_1 - w_2 \right \Vert_{L_2}^2 = \int_0^1 \int_0^1 \left( w_1(x,y) - w_2(x,y) \right)^2 \del x \, \del y. \label{eq:l2}
\end{equation}

\begin{proposition}[Graph distance is consistent]
\label{prop:grah_dt_conc}
Let $w_1$ and $w_2$ satisfy Assumptions~\ref{assum:lipschitz}--\ref{assum:equivalence}.
Let $G_1$ and $G_2$ be random graphs with at least $n$ nodes sampled from the graphons $w_1$ and $w_2$, respectively.
If $n\to\infty$ and $n_0$ is chosen such that
$\frac{n_0^2\log n}{n} \rightarrow 0$, then with high probability (w.h.p.),
\begin{align}
    \left \vert \left \Vert \,w_1 - w_2 \,\right \Vert_{L_2} - d(G_1,G_2) \right \vert &=  \mathcal{O} \textstyle \left( \frac{1}{n_0} \right).    \label{eq:A_w}
\end{align}
\end{proposition}
\textit{Proof sketch.}
We define a novel technique for approximating the graphon.
The proof in Appendix~\ref{sec:proof_prop1} first establishes that the approximation error is bounded using Assumption~\ref{assum:lipschitz}.
Consequently, a relation between approximated graphons and transformed graphs is derived using lemmas from \citet{chan2014consistent}.
Proposition~\ref{prop:grah_dt_conc} is subsequently proved using the above two results.
\hfill $\Box$

\textbf{Notation.}
For ease of exposition, Proposition~\ref{prop:grah_dt_conc} as well as main results are stated asymptotically using the standard $\mathcal{O}(\cdot)$ and $\Omega(\cdot)$ notations, which subsume absolute and Lipschitz constants.
We use ``with high probability'' (w.h.p.) to state that the probability of an event converges to $1$ as $n\to\infty$.

\section{Graph Clustering}
\label{sec:clustering}

We now present the first application of the proposed graph distance~\eqref{eq:graphdist} in the context of clustering network-valued data.
We are particularly interested in the setting where one needs to cluster a small population of large graphs, that is, minimum graph size $n$ grows faster than the sample size $m$. 
This scenario is relevant in practice as bioinformatics or neuroscience application often deals with very few graphs (see real datasets in Section \ref{sec:experiment}). Theoretically, this perspective complements guarantees for (graph) kernels that are applicable only in supervised setting and large sample regime, $m\to\infty$.
In contrast, our guarantees are more conclusive for bounded $m$ and large graph size, $n \rightarrow \infty$.

\textbf{Strategy for clustering.}
Since our aim is to cluster graphs of varying sizes, we transform the graphs to a common representation of $n_0\times n_0$ matrices, and use the graph distance function in~\eqref{eq:graphdist}.
We then use two different approaches for clustering: spectral clustering based on distances \citep{mukherjee2017clustering}, and similarity-based semi-definite programming \citep{perrot2020nearoptimal}.
We discuss the methods below, and prove statistical consistency, assuming that the graphs are sampled from graphons.

\subsection{Distance Based Spectral Clustering (DSC)}
Given $m$ graphs with adjacency matrices $G_1,...,G_m$, we propose a distance based clustering algorithm where we apply spectral clustering to an estimated distance matrix.
The distance matrix $\widehat{D} \in \mathbb{R}^{m\times m}$ is computed on all pairs of graphs using the defined estimator function~\eqref{eq:graphdist}, that is $\widehat{D}_{ij} = d(G_i,G_j)$. 
Unlike the standard Laplacian based spectral clustering, which is applicable for adjacency or similarity matrices, we use the method suggested by \citet{mukherjee2017clustering} that computes the $K$ leading eigenvectors  of $\widehat{D}$ (corresponding to the $K$ smallest eigenvalues in magnitude) and applies k-means clustering to the  rows of the eigenvector matrix resulting in $K$ number of clusters.
We refer to this distance based clustering algorithm as DSC, described in Algorithm~\ref{alg:spectral_clust} of Appendix. 
%
%
To derive the statistical consistency of DSC, we consider the problem of clustering $m$ random graphs of potentially different sizes, each sampled from one of $K$ graphons. 
We establish the consistency in Theorem \ref{theo:spect_error} by proving that the number of misclustered graphs goes to zero asymptotically (for large graphs).

\begin{theorem}[Consistency of DSC]
Consider $K$ graphons satisfying Assumptions~\ref{assum:lipschitz}--\ref{assum:equivalence}, and  $m$ random graphs $G_1,\ldots,G_m$, each sampled from one of the $K$ graphons (assume there is at least one graph from each graphon).
Define the distance matrix $D \in \mathbb{R}^{m\times m}$ such that $D_{ij} = \Vert w_i - w_j \Vert_{L_2}$ where $w_i$ and $w_j$ are the graphons from which $G_i$ and $G_j$ are generated.
Let $n$ be the size of the smallest graph, and $\gamma$ be the $K$-th smallest eigenvalue value of $D$ in magnitude.
As $n \rightarrow \infty$, if $n_0$ is chosen such that $\frac{m^2 n_0^2\log n}{n} \rightarrow 0$,
then DSC misclusters at most $
\mathcal{O}\left(\frac{m^3}{\gamma^2 n_0^2} \right)$ graphs w.h.p.
\label{theo:spect_error}
\end{theorem}

\textit{Proof sketch.} The proof, given in Appendix~\ref{sec:proof_dsc}, uses Davis-Kahan spectral perturbation theorem to bound the error in terms of $\Vert \widehat{D}-D\Vert_F$, which is further bounded using Proposition~\ref{prop:grah_dt_conc}.
\hfill $\Box$

While the number of misclustered graphs seem to depend on $m^3$,
we note that there is an inverse dependence on $\gamma^2$ which has dependence on $m$ (see Corollary 1 that illustrates it for a specific case).
Moreover, our focus is on the setting where $m=\mathcal{O}(1)$ and $n,n_0\to\infty$, in which case, the error asymptotically vanishes. It is natural to wonder whether the dependence on $m$ and $n_0$ is tight in the above bounds. Currently, we do not know the optimal rates, but deriving this would be difficult due to the strong dependency of entries in $\widehat{D}$ and slow rate of convergence of the graph distance in Proposition \ref{prop:grah_dt_conc}. 
The presence of $\gamma$ in the above clustering error bound makes Theorem \ref{theo:spect_error} less interpretable. 
Hence, we also consider the specific case of $K=2$ (two graphons) in the following result, along with the assumption that equal number of graphs are generated from both graphons.

\begin{corollary}
Let $w\neq w^\prime$ be two graphons satisfying Assumptions~\ref{assum:lipschitz}--\ref{assum:equivalence}, and $m$ is a bounded even number.
Assume that equal number of graphs are generated from $w$ and $w^\prime$. For any $n_0$ and large enough constant $C$ such that $\Vert w - w^\prime \Vert_{L_2} \geq C\frac{m}{n_0}$ and $\frac{m^2 n_0^2 \log n}{n} \rightarrow 0$ as $n \rightarrow \infty$, the number of  graphs misclustered by Algorithm~\ref{alg:spectral_clust} goes to zero w.h.p.
\label{theo:spect_k2}
\end{corollary}

The corollary implies that given the observed graphs are large enough, and if the choice of $n_0$ is relatively small, $n_0 \ll \sqrt{n/\log n}$, and the graphons are $\Omega(\frac{1}{n_0})$ apart in $L_2$-distance, then the clustering is consistent.
Intuitively, it can be understood that if we condense large graphs to a small representation (small $n_0$),  then the clusters can be identified only if the models are quite dissimilar.

\subsection{Similarity Based Semi-Definite Programming (SSDP)}
We propose another algorithm for clustering $m$ graphs based on similarity between pairs of graphs. 
The pairwise similarity matrix $\widehat{S} \in \mathbb{R}^{m\times m}$ is computed by applying Gaussian kernel on the distance between the graphs, 
that is $\widehat{S}_{ij} = \exp \left(-\frac{d(G_i,G_j)}{\sigma_i \sigma_j} \right)$, where $\sigma_1,\ldots,\sigma_n$ are parameters. 
For theoretical analysis, we assume $\sigma_1=\ldots=\sigma_n$ is fixed, but in experiments, the parameters are chosen adaptively. 
We use the following semi-definite program (SDP) \citep{yan2018provable,perrot2020nearoptimal} to find membership of the observed graphs.  
%
Let $X\in\sR^{m\times m}$ be the normalised clustering matrix, that is, $X_{ij} = {1}/{\vert \mathcal{C} \vert}$ if $i$ and $j$ belong to the same cluster $\mathcal{C}$, and $0$ otherwise. Then, the SDP for estimating $X$ is as follows: 
 \begin{align}
    &\max_X \, \text{trace}(\widehat{S}X) 
    \qquad\qquad\text{s.t.} \, X \geq 0, \,  X \succeq 0,  \, X\textbf{1} = \textbf{1}, \, \text{trace}(X) = K, \label{eq:sdp}
 \end{align}
 where $ X \geq 0, \,  X \succeq 0$ ensure that $X$ is a non-negative, positive semi-definite matrix, and $\textbf{1}$ denotes the vector of all ones. 
 We denote the optimal $X$ from the SDP  as $\widehat{X}$.
Once we have $\widehat{X}$, we apply standard spectral clustering on $\widehat{X}$ to obtain a clustering of the graphs.
We refer to this algorithm as SSDP, described in Algorithm~\ref{alg:sdp_clust} of Appendix.
%
%
We present strong consistency result for SSDP below.
\begin{theorem}[Consistency of SSDP]
\label{theo:sdp}
Consider $K$ graphons, $w_1,\ldots,w_K$, satisfying Assumptions~\ref{assum:lipschitz}--\ref{assum:equivalence}, and $m$ random graphs, each sampled from one of the $K$ graphons. Let $n$ be the size of the smallest graph.
As $n \rightarrow \infty$, if $n_0$ is chosen such that $\frac{m^2 n_0^2\log n}{n} \rightarrow 0$ and $\min\limits_{l \neq l^\prime} \Vert w_l - w_{l^\prime} \Vert_{L_2} = \Omega \left( \frac{m}{n_0}  \right)$, then 
%
the number of  graphs misclustered by SSDP is zero w.h.p.
\label{theo:sdp_sim}
\end{theorem}

\textit{Proof sketch.} The proof in Appendix~\ref{sec:proof_ssdp} adapts \citet[][Proposition 1]{perrot2020nearoptimal} to the present setting and combines it with Proposition~\ref{prop:grah_dt_conc} to derive the stated condition for zero error. \hfill $\Box$

Theorem \ref{theo:sdp} is slightly stronger than Theorem~\ref{theo:spect_error}, or Corollary~\ref{theo:spect_k2}, since SSDP achieve a zero clustering error for large enough graphs. This theoretical merit of SDP over spectral clustering is known in the statistics literature. 
Similar to Corollary \ref{theo:spect_k2}, the choice of $n_0$ is important such that it does not violate the minimum $L_2$-distance condition in the theorem to ensure consistency.

\textbf{Remark on the knowledge of $K$.}
Above discussions assume that the number of clusters $K$ is known, which is not necessarily the case in practice.
To tackle this issue, one can estimate $K$ using Elbow method \citep{thorndike1953belongs} or approach from \citet{perrot2020nearoptimal},and then use it as input in our algorithms, DSC and SSDP.
One can modify the SDP \eqref{eq:sdp} and Theorem \ref{theo:sdp} to the case where $K$ is adaptively estimated. 
However, we found the corresponding algorithm, adapted from \citet{perrot2020nearoptimal}, to be empirically unstable in the present context.
Hence, the knowledge of $K$ is assumed in the following experiments, which also allows the efficacy of the proposed algorithms and graph distance to be evaluated without the error induced by incorrect estimation of $K$.

\subsection{Experimental Analysis}
\label{sec:experiment}
In this section, we evaluate the performance of our algorithms DSC and SSDP, both in terms of accuracy and computational efficacy.
We measure the performance of the algorithms in terms of error rate, that is, the fraction of misclustered graphs by using the source of the graphs as labels. Since clustering provides labels up to permutation, we use the Hungarian method~\citep{kuhn1955hungarian} to match the labels.
The performance can also be measured in terms of Adjusted Rand Index (results in Appendix~\ref{sec:exp_ari}).
We use both simulated and real datasets for evaluation and obtain all our experimental results using Tesla K80 GPU instance with 12GB memory from Google Colab.

\textbf{Simulated data.}
We generate graphs of varied sizes from four  graphons, $W_1(u,v) = uv$, $W_2(u,v) = \exp \left\{ -\max ( u,v )^{0.75} \right\}$, $W_3(u,v) = \exp \left\{ - 0.5 * ( \min (u,v) + u^{0.5} + v^{0.5} )  \right\}$ and $W_4(u,v) = \vert u-v \vert$.
The simulated graphs are dense and the graph sizes are controlled to study how algorithms scale.
Their corresponding $L_2$ distances between pairs of graphons is shown later in Figure~\ref{fig:testing} and the heatmap of the graphons are visualised in Figure~\ref{fig:heatmap} in Appendix.

\begin{figure}[t]
\centering
\includegraphics[scale=0.8]{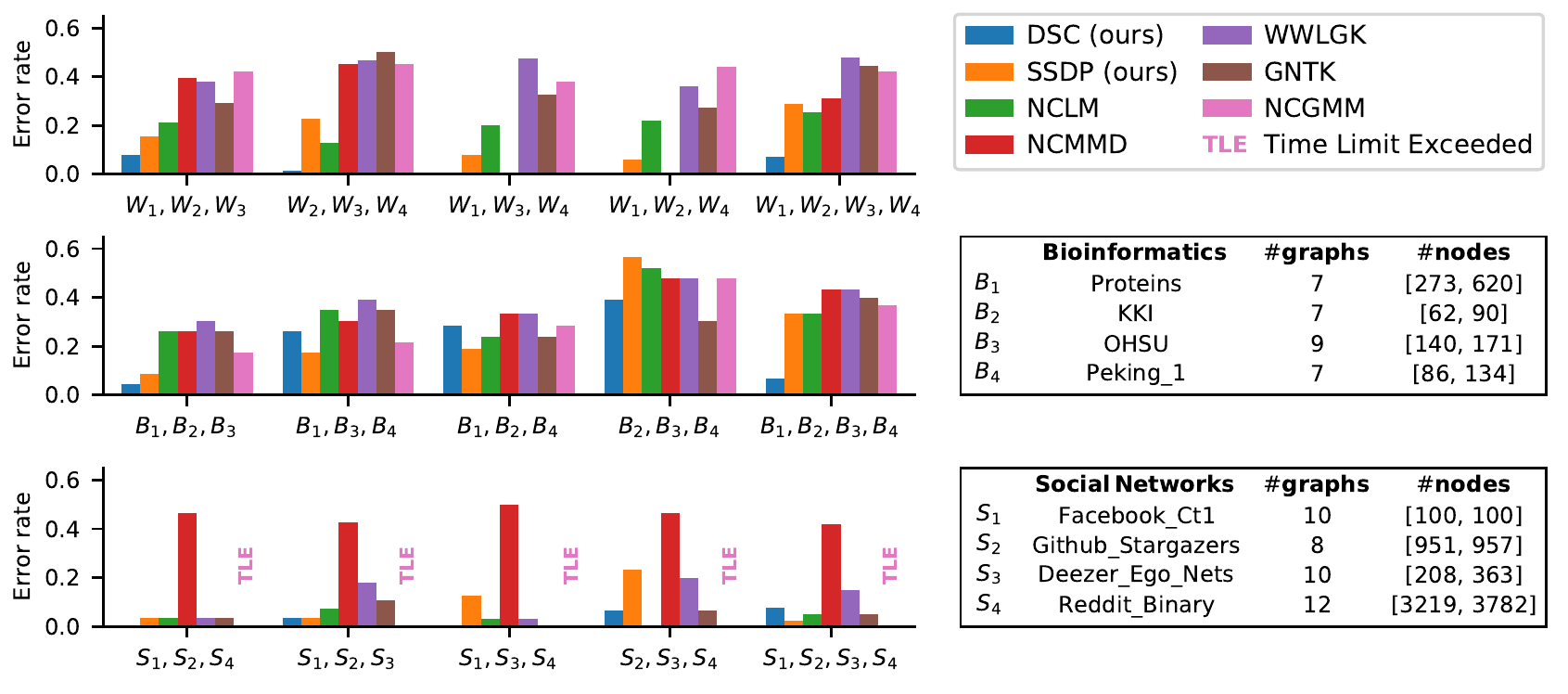}
\caption{Evaluation of DSC and SSDP with other methods. \textbf{(row 1)} Results on simulated data. \textbf{(rows 2 and 3)} Results on real data from Bioinformatics and Social Networks, respectively. DSC outperforms in majority of the cases. Tables in rows 2 and 3 show details of the considered datasets. }
\label{fig:real_data}
\end{figure}

\textbf{Real data.}
We analyse the performance of algorithms using datasets from two contrasting domains: molecule datasets from Bioinformatics and network datasets from Social Networks.
The Bioinformatics networks are smaller whereas the latter has relatively larger graphs.
We use \emph{Proteins} \citep{borgwardt2005protein}, \emph{KKI} \citep{pan2016task}, \emph{OHSU} \citep{pan2016task} and \emph{Peking\_1} \citep{pan2016task} datasets from Bioinformatics, and \emph{Facebook\_Ct1} \citep{oettershagen2020temporal}, \emph{Github\_Stargazers} \citep{rozemberczki2020api}, \emph{Deezer\_Ego\_Nets} \citep{rozemberczki2020api} and \emph{Reddit\_Binary} \citep{yanardag2015deep} datasets from Social Networks.
We sub-sample a few graphs from each dataset by setting a minimum number of nodes to validate the case of clustering small number of large graphs (small $m$, large $n$).
The number and size of the graphs sampled from each dataset are listed in ‘\#graphs’ and ‘\#nodes’ columns of tables in Figure~\ref{fig:real_data}.
We evaluate the clustering performance on all combinations of the datasets for three and four clusters in both the domains separately.


\textbf{Choice of $n_0$ and $\sigma_i$.} 
As noted in our algorithms DSC and SSDP, $n_0$ is an input parameter. 
Theorems~\ref{theo:spect_error} and \ref{theo:sdp_sim} show that the performance of both DSC and SSDP depend on the choice of $n_0 = \mathcal{O} \big( \sqrt{{n}/{\log n}}\big)$.
In the experiments, we set $n_0 = \sqrt{n/\log n}$ where $n$ is the minimum number of nodes.
In Appendix~\ref{sec:n0}, we use simulated data to  show that the above choice of $n_0$ is reasonable (if not the best) for both DSC and SSDP. 
Furthermore, the similarity matrix $\widehat{S}$ in SSDP is computed using parameters $\sigma_1, \ldots, \sigma_n$. 
In the experiments, we set $\sigma_i = d(G_i,G_{5nn})$ where $G_{5nn}$ is the fifth nearest neighbour of $G_i$.
Hence, apart from knowledge of $K$, our algorithms are parameter-free.

\textbf{Performance comparison with existing methods.}
We compare our algorithms with a range of approaches for measuring similarity or distance among multiple networks.
Most methods discussed below provide a kernel or distance matrix to which we apply spectral clustering to obtain the clusters: 

{1) Network Clustering based on Log-Moments (NCLM)} is the only known clustering strategy for graphs of different sizes \citep{mukherjee2017clustering}. It is based on network statistics called log moments.
Log moments for a graph with adjacency matrix $A$ and number of nodes $n$ is obtained by $\lp \log \lp m_1(A) \rp, \log \lp m_2(A) \rp, \ldots, \log \lp m_J(A) \rp \rp$ where $m_i(A) = \mathrm{trace}(A/n)^i$ and $J$ is a parameter.

{2) Wasserstein Weisfeiler-Lehman Graph Kernels (WWLGK)} is a recent graph kernel that is based on the Wasserstein distance between the node feature vector distributions of two graphs proposed by \citet{togninalli2019wasserstein}. 

{3) Graph Neural Tangent Kernel (GNTK)} is another graph kernel that describes infinitely wide graph neural networks derived by \citet{du2019graph}. Both WWLGK and GNTK provide state-of-the-art performance in graph classification with GNTK outperforming most graph neural networks. 

{4) Network Clustering algorithm based on Maximum Mean Discrepancy (NCMMD)} considers a graph metric (MMD) to cluster the graphs. MMD distance between random graphs is proposed as an efficient test statistic for random dot product graphs \citep{agterberg2020nonparametric}.
We compute MMD between the graphs that are represented by latent finite dimensional embedding called spectral adjacency embedding with the dimension $r$ as a parameter.

{5) In Network Clustering algorithm based on Graph Matching Metric (NCGMM)}, we match two graphs of different sizes by appending null nodes to the small graph as described in \citet{guo2019quotient} and compute Frobenius norm between the matched graphs as their distance. 
%
Although both the considered graph metrics (MMD and graph matching) are for different purposes, we evaluate their efficacy in the context of clustering. 

The different parameters to tune in the algorithms include $n_0$ and $\sigma_i$ in our algorithms DSC and SSDP, $J$ in NCLM, number of iterations ($\#itr$) to perform in WWLGK, number of layers ($\#layer$) in graph neural networks for GNTK, $r$ in NCMMD and none in NCGMM.
We fix $n_0$ in our algorithms using the theoretical bound and $\sigma_i$ is set adaptively as discussed, whereas we tune the parameters for other algorithms by grid search over a set of values.


\textbf{Evaluation on simulated data.}
We sample $10$ graphs of varied sizes between $50$ and $100$ nodes from each of the four graphons in Figure~\ref{fig:heatmap}, and evaluate the performances of all the seven algorithms.
We perform the experiments by considering all combinations of three and four clusters of the chosen graphons.
Based on the theoretical bound, $n_0$ is fixed to $5$ since minimum number of nodes is $50$. We report the performance for $J=8$, $r=3$, $\#itr=1$ and $\#layer=2$ as these produce the best results.
The first row of Figure~\ref{fig:real_data} shows the average performance of the algorithms computed over $5$ independent runs.
We observe that our algorithm DSC outperforms all the other algorithms, achieving nearly zero error in all cases, and SSDP also performs competitively by standing second or third best. 
The graph kernels, WWLGK and GNTK, and the graph metric based method NCGMM typically do not perform well. NCMMD either performs very well or quite poorly.
We sample small graphs since otherwise GNTK cannot run due to memory requirement for dense large graphs and NCGMM has high computation time. Appendix~\ref{sec:large_graph_exp} includes evaluation of the algorithms except GNTK and NCGMM on larger graphs, where we observe similar behaviour.

\textbf{Evaluation on real data.}
We consider all combinations of three and four clusters of both Bioinformatics and Social Networks separately, and evaluate the performance of the discussed seven algorithms. 
The second and third rows of Figure~\ref{fig:real_data} show the performance with $n_0=30, J=8$, $r=3$, $\#itr=1$ and $\#layer=2$, and the upper limit of $7200$ seconds ($2$ hours) as running time of algorithms.
We observe DSC outperforms other algorithms by a large margin in majority of the combinations, 
while in the other combinations like \{Proteins,KKI,Peking\_1\}, DSC performs well with a very small margin to the best performing one. 
Although NCLM and GNTK compare favorably in Social Networks datasets, they typically have high error rate in Bioinformatics datasets or simulated data, suggesting that they could be well suited for large networks, whereas DSC is more versatile and suitable for all networks.
The performance of SSDP is moderate on real data, but it achieves the smallest error in some cases, implying that SSDP is suited for certain types of networks.





\begin{wrapfigure}{r}{5cm}
\vspace{-15pt}
\includegraphics[width=5cm]{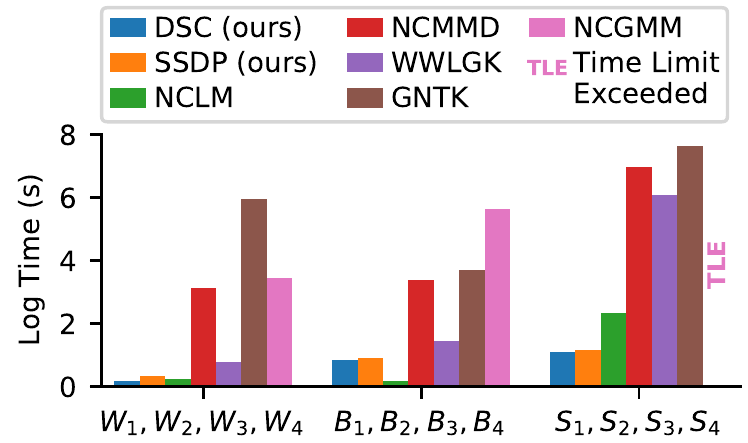}
\caption{Computation time}\label{fig:time_comp}
\end{wrapfigure}

\textbf{Computation time comparison.}
Figure~\ref{fig:time_comp} shows the time (measured in seconds) taken by each algorithm for four clusters case, plotted in log scale.
Similar behavior is observed in three clusters case also and the result can be found in Appendix~\ref{sec:time_all}.
Our algorithms, DSC and SSDP, perform competitively with respect to time as well.
In addition, it scales effectively for large graphs unlike other algorithms. It is worth noting that although NCLM takes lesser time than DSC and SSDP for small graphs, it takes longer for large social networks datasets, thus favoring our methods over NCLM in terms of both accuracy and scalability. 
Graph matching based algorithm, NCGMM, has severe scalability issue demonstrating the inapplicability of such methods to learning problems.
We also evaluate the scalability of the considered algorithms by measuring the time taken for clustering different sets of varied sized graphs from graphons $W_1,W_2,W_3$ and $W_4$.
Detailed discussion on the experiment is provided in Appendix~\ref{sec:scalability_exp}.
The experimental results also illustrate the high scalability of DSC and SSDP compared to the other algorithms.



\section{Graph Two-Sample Testing}
\label{sec:hyp_test}

Inspired by the remarkable performance of the proposed graph distance \eqref{eq:graphdist} in clustering, we analyse the applicability of the distance for graph two-sample testing.
Two-sample testing is usually studied in the large sample case $m\to\infty$, and several nonparametric tests are known that could also be applied to graphs.
However, in the context of graphs, it is relevant to study the small sample setting, particularly $m=2$, that is, the problem of deciding if two large graphs are statistically identical or not \citep{ghoshdastidar2020two,agterberg2020nonparametric}. 

We consider the following formulation of the graph two-sample problem, stated under the assumption that the graphs are sampled from graphons.
Given two random graphs, $G_1$ sampled from some model (here, graphon $w_1$), and $G_2$ from another model $w_2$, the goal is to determine which of the following  hypothesis is true: 
$H_0: \{ w_1=w_2 \}$ or $H_a: \left\{ w_1 \ne w_2 : \left\Vert w_1 - w_2 \right\Vert_{L_2} \geq \phi \right\}$
%
for some $\phi > 0$.
Existing works consider alternative random graph model, such as inhomogeneous Erd\H{o}s-R{\'e}nyi models or random dot product graph models, which are more restrictive.
The condition $\phi>0$ is necessary if one only has access to finitely many independent samples \citep{ghoshdastidar2020two}.
%
%
A two-sample test $T$ is a binary function of the given samples such that $T=1$ denotes that the test rejects the null hypothesis $H_0$ and $T=0$ implies that the test rejects the alternate hypothesis $H_a$.
The goodness of a two-sample test is measured in terms of the Type-I and Type-II errors, which denote the probabilities of incorrectly rejecting the null and alternate hypotheses, respectively.

The goal of this section is to show that one can construct a test $T$ that has arbitrarily small Type-I and Type-II errors.
For this purpose, we consider the test
\begin{equation}
\label{eq:test}
T: \mathbb{I}\left\{ d(G_1,G_2) \geq \xi \right\}
\end{equation}
for some $\xi>0$, where $\mathbb{I}\{\cdot\}$ is the indicator function and $d(G_1,G_2)$ is the proposed graph distance for some choice of integer $n_0$.
We state the following theoretical guarantee for the two-sample test $T$, where the performance is quantified in terms of Type-I and Type-II errors. 

\begin{theorem}
Assume that the graphons $w_1,w_2$ satisfy Assumptions~\ref{assum:lipschitz}--\ref{assum:equivalence}, and let the graphs $G_1\sim w_1$ and $G_2\sim w_2$ have at least $n$ nodes.
As $n\to\infty$, there is a choice of $\xi$ such that the Type-I and Type-II errors of the test $T$ in \eqref{eq:test} go to 0 if $\frac{n_0^2\log n}{n}\to0$ and $\phi \geq \frac{C}{n_0}$, where  the constant $C$ depends only on the Lipschitz constants.
\label{theo:hyp_test}
\end{theorem}

Theorem \ref{theo:hyp_test} shows that the test $T$ in \eqref{eq:test} can distinguish between any pair of graphons that have separation $\Vert w_1 - w_2\Vert_{L_2} = \Omega(1/n_0)$ with arbitrarily small error, if the graphs are large enough.

\begin{figure}[t]
\centering
\includegraphics[scale=0.75]{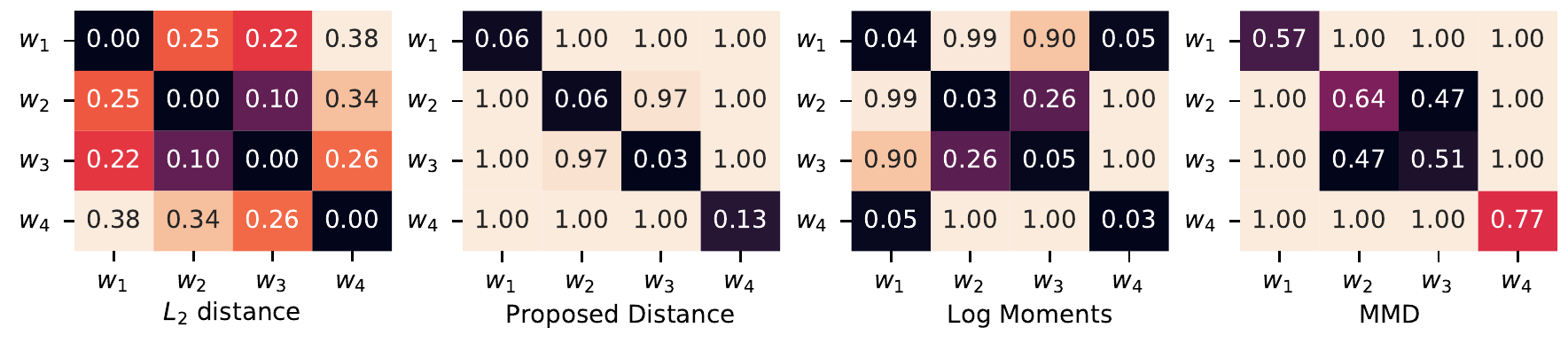}
\caption{\textbf{(left)} $L_2$ distance between the graphons $W_1$, $W_2$, $W_3$ and $W_4$. \textbf{(other plots)} Average power of the test \eqref{eq:test} for a graph pair of sizes $100$ and $200$, sampled from every pair of graphons.}
\label{fig:testing}
\end{figure}

\textbf{Empirical analysis.}
We empirically validate the consistency result in Theorem~\ref{theo:hyp_test} by computing power of the proposed two-sample test $T$, which measures the probability of rejecting the null hypothesis $H_0$.
Intuitively, power of the test for graphs sampled from same graphons should be small (close to a pre-specified significance level) since $H_0$ must not be rejected, whereas, it should be close to $1$ for graphs sampled from different graphons.
As known in the testing literature, theoretical threshold, $\xi$ in \eqref{eq:test}, is typically conservative in practice and the rejection/acceptance is decided based on $p$-values, computed using bootstrap samples.
To this end, we follow the bootstrapping strategy in~\citet[Boot-ASE algorithm]{ghoshdastidar2018practical}.
In addition, we compare the proposed test $T$ by replacing $d(G_1,G_2)$ in \ref{eq:test} with two other statistics, log moments from \citet{mukherjee2017clustering} and MMD, an efficient test statistics for random dot product graphs~\citep{agterberg2020nonparametric}.
We perform the experiment by sampling two graphs $G_1 \sim w_1$ and $G_2 \sim w_2$ of size $n$ and $2n$, respectively, where $w_1$ and $w_2$ are chosen from the graphons $W_1, W_2, W_3, W_4$ discussed in Section~\ref{sec:experiment}.
We consider $n=100$ and thus $n_0=10$ from the theoretical bound using $n$ for evaluating the test $T$. 
The power of test $T$ is computed for the significance level $0.05$, averaged over $500$ trials of bootstrapping $100$ samples generated from all pairs of graphons.
The plots in Figure~\ref{fig:testing} show the average power of test $T$ with our proposed distance, log moments and MMD as $d(G_1,G_2)$, respectively.
From the results, it is clear that test $T$ using our proposed distance can distinguish between pairs of graphons that are quite close too, for instance, $W_2$ and $W_3$ (smallest $L_2$ distance), whereas, other test statistics are weak as log moments statistic accepts the null hypothesis even when it is wrong (see $W_1$ and $W_4$) and MMD based test rejects it strongly almost always (see diagonal).
In Appendix~\ref{app:exp}, we present the results for smaller $n (50)$ and larger $n (150)$ with similar observations and evaluation of test $T$ on real datasets where our proposed distance achieves the best results.

\section{Conclusion}
\label{sec:conclusion}

There has been significant progress in learning on complex data, including network-valued data.
However, much of the theoretical and algorithmic development have been in large sample problems, where one has access to $m\to\infty$ independent samples.
Practical applications of network-valued data analysis often leads to small sample, large graph problems---a setting where the machine learning literature is quite limited.
Inspired by graph limits and high-dimensional statistics, this paper proposes a simple graph distance \eqref{eq:graphdist} based on non-parametric graph models (graphons). 

Sections \ref{sec:clustering}--\ref{sec:hyp_test} demonstrate that the proposed graph distance leads to provable and practically effective algorithms for clustering (DSC and SSDP) as well as two-sample testing \eqref{eq:test}.
Extensive empirical studies on simulated and real data show that the clustering based on the graph distance \eqref{eq:graphdist} outperforms methods based on more complex graph similarities or metrics, both in terms of accuracy and scalability.
Figures \ref{fig:real_data}--\ref{fig:time_comp} show that DSC achieves best performance for both small dense graphs (simulated graphons) as well as large sparse graphs (social networks).
On the other hand, popular machine learning approaches---graph kernels or graph matching---can be computationally expensive in large graphs and their performance may not improve as $n\to\infty$, see WWLGK in Figure \ref{fig:sim_data_large}.

Statistical approaches, such as the proposed clustering algorithms and two-sample test, show better performance on large graphs (Figures \ref{fig:real_data}, \ref{fig:testing} and Appendix \ref{sec:large_graph_exp}).
Theorems~\ref{theo:spect_error}--\ref{theo:hyp_test} theoretically support this observation by showing consistency of the clustering and testing methods in the limit of $n\to\infty$.
The theoretical results, however, hinge on Assumptions \ref{assum:lipschitz}--\ref{assum:equivalence}.
We remark that such smoothness and equivalence assumptions could be necessary for meaningful non-parametric approaches, which is also supported by the graph testing and graphon estimation literature. Further insights about the necessity of smoothness assumptions would aid in theoretical and algorithmic development.

The poor performance of graph kernels and graph matching in clustering and small sample problems calls for further studies on these methods, which have shown success in network classification.
Fundamental research, combining graphon based approaches and kernels, could lead to improved techniques.
Algorithmic modifications, such as estimation of $K$, would be also useful in practice.

\section{Acknowledgment}
This work has been supported by the German Research Foundation (Research Training Group GRK 2428) and the Baden-W{"u}rttemberg Stiftung (Eliteprogram for Postdocs project ``Clustering large evolving networks'').
The authors thank the International Max Planck Research School for Intelligent Systems (IMPRS-IS) for supporting Leena Chennuru Vankadara.

 \clearpage


\clearpage
\bibliography{ref}
\bibliographystyle{iclr2022_conference}

\clearpage
\section{Proofs of theoretical results} \label{sec:proofs}
We discuss the proofs of Proposition \ref{prop:grah_dt_conc} and Theorems~\ref{theo:spect_error}--\ref{theo:hyp_test} with supporting lemmas in this section.

\subsection{Proposition~\ref{prop:grah_dt_conc}}
\label{sec:proof_prop1}
The distance function defined in \eqref{eq:graphdist} estimates the $L_2$-distance between graphons that are continuous. 
To prove this, we introduce a method to discretize the continuous graphons in the following so that it is comparable with the transformed graphs described in the graph distance estimator \eqref{eq:graphdist}. 

\textbf{Graphon discretization.}
We discretize the graphon by applying piece-wise constant function approximation that is inspired from \citet{chan2014consistent}, similar to the graph transformation. More precisely, any continuous graphon $w$ is discretized to a matrix $W$ of size $n_0\times n_0$ with
\begin{align}
    W_{ij} = \frac{1}{1/n_0^2}\int_{0}^{\frac{1}{n_0}} \int_{0}^{\frac{1}{n_0}} w(x+\frac{i}{n_0}, y+\frac{j}{n_0}) \,\del x \,\del y \label{eq:graphon_apprx}
\end{align}

We make Assumptions \ref{assum:lipschitz}--\ref{assum:equivalence} to derive the concentration bound stated in Proposition \ref{prop:grah_dt_conc}. 
The proof structure is as follows:
\begin{enumerate}
    \item We bound the point-wise deviation of graphon with its discretized correspondence using Lipschitzness assumption (Lemma \ref{lemma:lipschitz}).
    \item We derive the error bound between the $L_2$-distance and Frobenius norm of the discretized graphons using Lemma \ref{lemma:lipschitz} (Lemma \ref{lemma:err_bound}).
    \item We establish a relation between Frobenius norm of the histograms of  graphons and graphs (Lemma \ref{lemma:A_W}).
    \item Finally, we prove Proposition \ref{prop:grah_dt_conc} by combining Lemmas \ref{lemma:err_bound} and \ref{lemma:A_W}.
\end{enumerate}

\begin{lemma}[Lipschitz condition] 
For any graphon $w$ and corresponding discretization $W$, define a piecewise constant function $\overline{w(x,y)}= W_{ij}$ where $x \in [\frac{i}{n_0}, \frac{i}{n_0}+\frac{1}{n_0}]$ and $y \in [\frac{j}{n_0}, \frac{j}{n_0}+\frac{1}{n_0}]$. 
Using the Lipschitz continuous assumption, we have
\begin{align}
    \left\vert \overline{w(x,y)} -w(x,y) \right\vert \leq \frac{2 \sqrt{2} L}{n_0}, \label{eq:lipschitz}
\end{align}
where $L$ is the Lipschitz constant in Assumption~\ref{assum:lipschitz}.
\label{lemma:lipschitz}
\end{lemma}

\textbf{Proof of Lemma \ref{lemma:lipschitz} (Lipschitz condition).} 
We use Assumption \ref{assum:lipschitz} on Lipschitzness to prove this lemma. The following holds for a graphon $w$ with Lipschitz constant $L$, 
\begin{equation}
     \left\vert w(x+\frac{i}{n_0},y+\frac{j}{n_0}) -w(x,y) \right\vert \leq L \sqrt{\frac{i^2}{n_0^2} + \frac{j^2}{n_0^2}} \leq \frac{L\sqrt{2}}{n_0} \label{eq:lip_}
\end{equation}
We prove Lemma~\ref{lemma:lipschitz} using \eqref{eq:lip_} and the definition of $\overline{w(x,y)} = W_{ij}$ where $x \in [\frac{i}{n_0}, \frac{i}{n_0}+\frac{1}{n_0}]$ and $y \in [\frac{j}{n_0}, \frac{j}{n_0}+\frac{1}{n_0}]$, 
\begin{align}
    \left\vert \overline{w(x,y)} -w(x,y) \right\vert &= \left\vert \overline{w(x,y)} -w(x,y) \pm w(x+\frac{i}{n_0},y+\frac{j}{n_0}) \right\vert \nonumber \\
    &\leq \left\vert \overline{w(x,y)} - w(x+\frac{i}{n_0},y+\frac{j}{n_0}) \right\vert + \left\vert w(x+\frac{i}{n_0},y+\frac{j}{n_0})  -w(x,y) \right\vert = \frac{2 \sqrt{2} L}{n_0} \square \nonumber
\end{align}

\begin{lemma}[Error bound of discretization]
For two graphons $w_1$ and $w_2$, the error bound between the $L_2$-distance and the Frobenius norm of the corresponding discretized graphons $W_1$ and $W_2$ satisfies
\begin{equation}
   \left \vert \Vert \,w_1 - w_2 \,\Vert_{L_2} - \dfrac{1}{n_0} \Vert W_1 - W_2 \Vert_F \right \vert \leq \frac{4\sqrt{2}L}{n_0} . \label{eq:appr_err}
\end{equation}
\label{lemma:err_bound}
\end{lemma}

\textbf{Proof of  Lemma \ref{lemma:err_bound} (Error bound of the approximation).} 
Lemma \ref{lemma:lipschitz} is used to prove this lemma. Let $L_1$ and $L_2$ be the Lipschitz constants of $w_1$ and $w_2$.
\begin{align}
    \left \Vert \,w_1 - w_2 \,\right \Vert_{L_2}^2 &= \int_0^1 \int_0^1 \left( w_1(x,y) - w_2(x,y) \right)^2 \,\del x \,\del y \nonumber \\
    \text{$\pm \overline{w_1(x,y)}$ } &\text{ and  $\pm \overline{w_2(x,y)}$ within the square, expand and apply Lipschitzness condition from Lemma~\ref{lemma:lipschitz}} \nonumber \\
    &\leq \frac{8L_1^2}{n_0^2} + \frac{8L_2^2}{n_0^2} + \frac{16L_1 L_2}{n_0^2} + \sum_{k=0}^{n_0} \sum_{l=0}^{n_0} \dfrac{1}{n_0^2} \Big( (W_1)_{kl} - (W_2)_{kl} \Big)^2 \nonumber \\
    &\qquad \qquad \qquad + \frac{4\sqrt{2}(L_1+L_2)}{n_0} \sum_{k=0}^{n_0} \sum_{l=0}^{n_0} \dfrac{1}{n_0^2} \Big \vert (W_1)_{kl} - (W_2)_{kl} \Big \vert  \nonumber \\
    &\leq \Big( \frac{2\sqrt{2}(L_1+L_2)}{n_0} \Big)^2 + \dfrac{1}{n_0^2} \Vert W_1  - W_2 \Vert_F^2 +  \frac{4\sqrt{2}(L_1+L_2)}{n_0^3} \Vert W_1 - W_2 \Vert_1 \nonumber \\
    &\stackrel{(a)}{\leq} \Big( \frac{2\sqrt{2}(L_1+L_2)}{n_0} + \dfrac{1}{n_0} \Vert W_1 - W_2 \Vert_F \Big)^2 \label{eq:upper_bound_err} 
\end{align}
$(a): \Vert x \Vert_1 \leq \sqrt{n} \, \Vert x\Vert_F $ 

Similarly, by $\pm w_1(x,y)$ and $\pm w_2(x,y)$ to $\dfrac{1}{n_0^2} \Vert W_1  - W_2 \Vert_F^2$ and applying Lipschitzness condition from Lemma~\ref{lemma:lipschitz} we get,
\begin{align}
    \dfrac{1}{n_0^2} \Vert W_1  - W_2 \Vert_F^2 &\leq \Big( \frac{2\sqrt{2}(L_1+L_2)}{n_0} + \Vert w_1 - w_2 \Vert_{L_2} \Big)^2 \label{eq:lower_bound_err}
\end{align}
Combining \eqref{eq:upper_bound_err} and \eqref{eq:lower_bound_err}, we prove
\begin{align}
    \left \vert \Vert \,w_1 - w_2 \, \Vert_{L_2} - \dfrac{1}{n_0} \Vert W_1 - W_2 \Vert_F \right \vert \leq  \frac{2\sqrt{2}(L_1+L_2)}{n_0} \stackrel{(b)}{\leq} \frac{4\sqrt{2}L}{n_0}, \quad \text{(b): $L = \max \{ L_1, L_2 \}$} \square  \nonumber
\end{align}

We derive the following relation between histogram of graphs and graphons by adapting lemmas from \citet{chan2014consistent} for our problem and the error bound of discretization \eqref{eq:appr_err}.
\begin{lemma} 
Let $G_1\sim w_1$  and $G_2\sim w_2$ have respective graph transformations $A_1$ and $A_2$.
Let $W_1$ and $W_2$ be the corresponding discretized graphons of $w_1$ and $w_2$, respectively.
As $n \rightarrow \infty$ and $\dfrac{n_0^2\log n}{n} \rightarrow 0$, then for any $\epsilon > 0$, 
\begin{align}
    \left \vert \Vert \,A_1 - A_2 \,\Vert_F - \Vert \,W_1 - W_2 \,\Vert_F \right \vert &\leq  4 \, \epsilon \label{eq:A_W}
\end{align}
with probability converging to $1$.
\label{lemma:A_W}
\end{lemma}

\textbf{Proof of Lemma \ref{lemma:A_W}.}
The proof of this lemma is inspired from \citet{chan2014consistent}. 
For $i=\{1,2\}$, let matrix $W_i$ of size $n_0 \times n_0$ be the discretized graphon $w_i$ and let matrix $\widehat{A}_i$ of size $n_0\times n_0$ be another transformation of graph $G_i$ based on the true permutation $\widehat{\sigma}_i$, that is, 
$\widehat{\sigma}_i$ denotes the ordering of the graph $G_i$ based on the corresponding graphon $w_i$. 
In other words, the discretized graphon $W_i$ is the expectation of $\widehat{A}_i$.  The reordered graph is denoted by $G_i^{\widehat{\sigma}}$ and then $\widehat{A}_i$ is obtained by,
\begin{align}
    (\widehat{A}_i)_{kl} = \frac{1}{h^2}\sum_{k_1=0}^{h} \sum_{l_1=0}^{h} (G^{\widehat{\sigma}_i})_{kh+k_1, lh+l_1} \quad \text{where $h = \lfloor \frac{n}{n_0} \rfloor$ and $\lfloor \cdot \rfloor$ is the floor function.} \nonumber 
\end{align}

We bound $\Vert A_1 - A_2 \Vert_F$ using $\widehat{A}_1$, $\widehat{A}_2$, $W_1$ and $W_2$ as,
\begin{align}
    \Vert A_1 - A_2 \Vert_F &\leq \Vert A_1 - \widehat{A}_1 \Vert_F + \Vert A_2 - \widehat{A}_2 \Vert_F + \Vert \widehat{A}_1 - \widehat{A}_2 \Vert_F  \nonumber \\ 
    &\leq 2 \max\limits_{i=\{1,2\}} \Vert A_i - \widehat{A}_i \Vert_F + \Vert \widehat{A}_1 - \widehat{A}_2 \Vert_F  \nonumber \\
    &\leq 2 \max\limits_{i=\{1,2\}} \Vert A_i - \widehat{A}_i \Vert_F + 2 \max\limits_{i=\{1,2\}}  \Vert \widehat{A}_i - W_i \Vert_F + \Vert W_1 - W_2 \Vert_F \label{eq:diff_bound_1}
\end{align}

We have the following for all $i$ using Assumption \ref{assum:lipschitz-deg} on Lipschitzness of the degree distribution $g(u)_i$ of graphon $w_i$ and Lemma 3 of \citet{chan2014consistent},
\begin{align}
    \mathbb{E}[\Vert A_i - \widehat{A}_i \Vert_F^2] &\leq \dfrac{n_0^4}{n_i^2}\left( 2+ 4C_i^2L_i^2\dfrac{\log n_i}{n_i}\right) + n_0^2\left( 4C_i^2L_i^2\dfrac{\log n_i}{n_i} \right) \,\, \text{; $C_i$ depends on Lipschitz constants of $g(u)_i$} \nonumber \\
    &\leq 2\dfrac{n_0^4}{n^2} + \dfrac{n_0^4}{n^2}4C_i^2L_i^2\dfrac{\log n}{n} + n_0^2 4C_i^2L_i^2\dfrac{\log n}{n} = \mathcal{O} \left( n_0^2\dfrac{\log n}{n} \right) \label{eq:A'_A}
\end{align}
Applying Markov's inequality to bound the probability $\mathbb{P}\left(\max\limits_{i=\{1,2\}} \Vert  A_i - \widehat{A}_i \Vert_F^2 \geq \epsilon_1^2 \right)$ using \eqref{eq:A'_A},
\begin{align}
    \mathbb{P}\left(\max\limits_{i=\{1,2\}} \Vert  A_i - \widehat{A}_i \Vert_F^2 \geq \epsilon_1^2 \right) &\stackrel{(a)}{\leq} \sum\limits_{i=\{1,2\}} \frac{\mathbb{E}\left[ \Vert  A_i - \widehat{A}_i \Vert_F^2  \right]}{\epsilon_1^2} \stackrel{(\ref{eq:A'_A})}{=} \mathcal{O} \left( \dfrac{n_0^2\log n}{ \epsilon_1^2 n} \right)  \,\, \text{; $(a):$ Union bound} \label{eq:pA'_A}
\end{align}
Thus asymptotically, as $n \rightarrow \infty$ and $\dfrac{n_0^2\log n}{ n} \rightarrow 0$, for  all $i$ and any $\epsilon_1 > 0$, $\Vert A_i - \widehat{A}_i \Vert_F < \epsilon_1$ with probability converging to $1$.

From Lemma 4 of \citet{chan2014consistent}, we have the following for all $i$,
\begin{equation}
    \mathbb{E}[\Vert \widehat{A}_i - W_i \Vert_F^2] \leq \frac{n_0^4}{n_i^2} \leq \frac{n_0^4}{n^2} \label{eq:EA_W}
\end{equation}
Applying Markov's inequality to bound the probability $\mathbb{P}\left(\max\limits_{i=\{1,2\}} \Vert  \widehat{A}_i - W_i \Vert_F^2 \geq \epsilon_2^2 \right)$ using \eqref{eq:EA_W},
\begin{align}
    \mathbb{P}\left(\max\limits_{i=\{1,2\}} \Vert  \widehat{A}_i - W_i \Vert_F^2 \geq \epsilon_2^2 \right) &\stackrel{(a)}{\leq} \sum\limits_{i=\{1,2\}} \frac{\mathbb{E}\left[ \Vert  \widehat{A}_i - W_i \Vert_F^2 \right]}{\epsilon_2^2} \stackrel{(\ref{eq:EA_W})}{\leq} \dfrac{2n_0^4}{\epsilon_2^2 n^2} && \text{; $(a):$ Union bound} \label{eq:pA_W}
\end{align}
Again asymptotically as $n \rightarrow \infty$, for  all $i$ and any $\epsilon_2 > 0$, $\Vert \widehat{A}_i - W_i \Vert_F < \epsilon_2$ with probability converging to $1$.
Lets assume $\epsilon_2 = \epsilon_1 = \epsilon$. Substituting \eqref{eq:pA'_A} and \eqref{eq:pA_W} in \eqref{eq:diff_bound_1},
\begin{align}
    \Vert A_1 - A_2 \Vert_F &\leq  4 \epsilon + \Vert W_1 - W_2 \Vert_F \label{eq:upper_bound}
\end{align}
with probability converging to $1$ as $n \rightarrow \infty$ and $\dfrac{n_0^2\log n}{ n} \rightarrow 0$.

The lower bound can similarly be obtained, 
\begin{align}
    \Vert A_1 - A_2 \Vert_F &\geq \Vert W_1 - W_2 \Vert_F - 2 \max\limits_{i=\{1,2\}} \Vert \widehat{A}_i - W_i \Vert_F - 2 \max\limits_{i=\{1,2\}} \Vert \widehat{A}_i - A_i \Vert_F  \nonumber \\
    &\geq \Vert W_1 - W_2 \Vert_F - 4 \epsilon \label{eq:lower_bound}
\end{align}
with probability converging to $1$ as $n \rightarrow \infty$ and $\dfrac{n_0^2\log n}{ n} \rightarrow 0$. \\
Thus, $\left\vert \Vert A_1 - A_2 \Vert_F -  \Vert W_1 - W_2 \Vert_F \right\vert \leq  4 \epsilon $ satisfy for any $\epsilon > 0$ with probability converging to $1$ as $n \rightarrow \infty$ and $\dfrac{n_0^2\log n}{ n} \rightarrow 0$  from equations \eqref{eq:upper_bound} and \eqref{eq:lower_bound}. $\hfill \square$

\textbf{Proof of Proposition \ref{prop:grah_dt_conc} (Graph distance is consistent).}
Proposition \ref{prop:grah_dt_conc} immediately follows from Lemmas \ref{lemma:err_bound} and \ref{lemma:A_W} after a simple decomposition step as shown below.
\begin{align}
    \left \vert \Vert \,w_1 - w_2 \, \Vert_{L_2} - d(G_1, G_2) \right \vert &= \left \vert \Vert \,w_1 - w_2 \, \Vert_{L_2} - d(G_1, G_2) \pm \dfrac{1}{n_0} \Vert W_1 - W_2 \Vert_F \right \vert \nonumber \\
    &\leq \left \vert \Vert \,w_1 - w_2 \, \Vert_{L_2} - \dfrac{1}{n_0} \Vert W_1 - W_2 \Vert_F \right \vert + \left \vert \dfrac{1}{n_0} \Vert W_1 - W_2 \Vert_F  - d(G_1, G_2) \right \vert  \nonumber \\
    &\stackrel{\ref{lemma:err_bound}, \ref{lemma:A_W}}{\leq} \frac{4\sqrt{2}L}{n_0} + \frac{4\epsilon}{n_0} = \mathcal{O} \left(  \frac{1}{n_0} \right) \text{holds for any $\epsilon>0$ as $n \rightarrow \infty$ and $\dfrac{n_0^2\log n}{ n} \rightarrow 0$ .} \square \nonumber
\end{align}

\subsection{Distance Based Spectral Clustering (DSC)}
\label{sec:proof_dsc}
We make Assumptions \ref{assum:lipschitz}--\ref{assum:equivalence} on the $K$ graphons 
to analyse the Algorithm~\ref{alg:spectral_clust}.
We establish the consistency of this algorithm by deriving the number of misclustered graphs $| \mathcal{M} |$ through the following steps.
\begin{enumerate}
    \item We establish deviation bound between the estimated distance matrix $\widehat{D}$ and the ideal distance matrix $D$ (Lemma \ref{lemma:D_hat_D}).
    \item We formulate Davis-Kahan theorem in terms of the deviation bound using the result from \citet{mukherjee2017clustering} (Lemma \ref{lemma:ortho_mat}).
    \item We derive the number of misclustered graphs from Lemma \ref{lemma:ortho_mat}.
\end{enumerate}

As stated previously, $\widehat{D}$ in Algorithm \ref{alg:spectral_clust} is an estimate of $D \in \mathbb{R}^{m\times m}$, where we define $D_{ij} = \Vert w_i - w_j \Vert_{L_2}$. Note that $D$ is a block matrix with rank $K$, since $D_{ij} = 0$ for all $G_i, G_j$ generated from same graphon, and equals the distance between the graphons $i$ and $j$ otherwise.

We derive the deviation bound for the distance matrix using Lemma \ref{lemma:A_W} and the result is as follows.
\begin{lemma}[Distance deviation bound]
As $n \rightarrow \infty$ and $\dfrac{n_0^2\log n}{n} \rightarrow 0$, we establish 
\begin{align}
    \left \Vert \widehat{D} - D \right \Vert_F = \mathcal{O} \left( \dfrac{m}{n_0} \right) \label{eq:spect_dist} 
\end{align}
with probability converging to $1$.
\label{lemma:D_hat_D}
\end{lemma}

\textbf{Proof of Lemma \ref{lemma:D_hat_D} (Distance deviation bound).}
From Proposition \ref{prop:grah_dt_conc} and the definitions of $\widehat{D}_{ij}$ and $D_{ij}$,  it is easy to see that $\left \vert  \widehat{D}_{ij} - D_{ij} \right \vert = \mathcal{O} \left( \dfrac{1}{n_0}\right)$ with probability converging to 1  as $n \rightarrow \infty$ and $\dfrac{n_0^2\log n}{ n} \rightarrow 0$.

Using Lemmas \ref{lemma:err_bound} and \ref{lemma:A_W}, and the definitions of $\widehat{D}_{ij}$ and $D_{ij}$, we have
\begin{align}
    \left \vert \widehat{D}_{ij} - D_{ij} \right \vert 
    &= \left \vert \dfrac{1}{n_0}\Vert A_i - A_j \Vert_F - \Vert w_i - w_j \Vert_{L_2} \right \vert \nonumber \\
    &= \left \vert \dfrac{1}{n_0}\Vert A_i - A_j \Vert_F \pm \dfrac{1}{n_0}\Vert W_i - W_j \Vert_F - \Vert w_i - w_j \Vert_{L_2} \right \vert \nonumber \\
    &\leq \dfrac{1}{n_0} \left \vert \Vert A_i - A_j \Vert_F - \Vert W_i - W_j \Vert_F \right \vert + \left \vert \Vert w_i - w_j \Vert_{L_2} - \dfrac{1}{n_0} \Vert W_i - W_j \Vert_F  \right \vert \nonumber \\
    &\stackrel{(\ref{eq:A_W})}{\leq} \dfrac{4 \epsilon}{n_0} + \dfrac{4\sqrt{2}L}{n_0} \quad \quad \text{with prob. $\rightarrow 1$ asymptotically} \nonumber  
\end{align}
Thus asymptotically, $\left \vert  \widehat{D}_{ij} - D_{ij} \right \vert = \mathcal{O} \left( \dfrac{1}{n_0}\right)$ with probability converging to 1.

Hence, $ \Vert \widehat{D} - D \Vert_F = \mathcal{O} \left( \dfrac{m}{n_0}\right)$ with probability converging to 1 as $n \rightarrow \infty$ and $\dfrac{m^2 n_0^2\log n}{ n} \rightarrow 0$. 

A variant of Davis-Kahan theorem~\citep{mukherjee2017clustering} and the derived deviation bound \eqref{eq:spect_dist} are used to prove the following lemma. 
\begin{lemma}[Davis-Kahan theorem]
Let $V$ and $\widehat{V}$ be the $m\times K$ matrices whose columns correspond to the leading $K$ eigenvectors of $D$ and $\widehat{D}$, respectively. Let $\gamma$ be the $K$-th smallest eigenvalue value of $D$ in magnitude. As $n \rightarrow \infty$ and $\dfrac{n_0^2\log n}{n} \rightarrow 0$, there exists an orthogonal matrix $\widehat{O}$ such that, 
\begin{align}
    \left \Vert \widehat{V} \widehat{O} - V \right \Vert_F = \mathcal{O} \left( \dfrac{ m}{\gamma n_0} \right) \label{eq:spect}
\end{align}
with probability converging to $1$. 
\label{lemma:ortho_mat}
\end{lemma}

\textbf{Proof of Lemma \ref{lemma:ortho_mat} (Davis-Kahan theorem).}
A variant of Davis Kahan theorem from Proposition A.2 of \citet{mukherjee2017clustering} states the following for matrix $D$ of rank $K$. Let $\widehat{V}$ and $V$ be $m*K$ matrices whose columns correspond to the leading $K$ eigenvectors of $\widehat{D}$ and $D$, respectively, and $\gamma$ be the $K$-th smallest eignenvalue of $D$ in magnitude, then there exists an orthogonal matrix $\widehat{O}$ of size $K*K$ such that, 
\begin{align}
    \left \Vert \widehat{V} \widehat{O} - V \right \Vert_F \leq \dfrac{4 \left \Vert \widehat{D} - D \right \Vert_F}{\gamma} \stackrel{\ref{lemma:D_hat_D}}{=} \mathcal{O} \left( \dfrac{m}{\gamma n_0}\right) \quad \text{as $n \rightarrow \infty$ and $\dfrac{m^2 n_0^2\log n}{ n} \rightarrow 0$.} \square \nonumber
\end{align}

The number of misclustered graphs is $| \mathcal{M} | \leq 8 m_T  \Vert \widehat{V} \widehat{O} - V \Vert_F^2$ where $m_T$ is the maximum number of graphs generated from a single graphon~\citep{mukherjee2017clustering}. 
Since $m_T=\mathcal{O}(m)$, $| \mathcal{M} | = \mathcal{O} \left( \dfrac{ m^3}{\gamma^2 n_0^2} \right)$ by substituting \eqref{eq:spect} in $| \mathcal{M} |$.
Hence proving Theorem~\ref{theo:spect_error}.

\textbf{Proof of Theorem \ref{theo:spect_error}.}
The number of misclustered graphs $\vert \mathcal{M} \vert \leq 8 m_T \left \Vert \widehat{V} \widehat{O} - V \right \Vert_F^2$ from \citet{mukherjee2017clustering}.
Thus, we prove the theorem using Lemma \ref{lemma:ortho_mat}. That is, as $n \rightarrow \infty$ and $\dfrac{m^2n_0^2\log n}{ n} \rightarrow 0$, 
\begin{align}
    \vert \mathcal{M} \vert \leq 8 m_T \left \Vert \widehat{V} \widehat{O} - V \right \Vert_F^2 \stackrel{\ref{lemma:ortho_mat}}{=} \mathcal{O} \left(\dfrac{m^3}{\gamma^2 n_0^2} \right) 
    \square \nonumber
\end{align}

\textbf{Proof of Corollary \ref{theo:spect_k2}.}
This corollary deals with a special case where $K=2$ and equal number of graphs are generated from the two graphons $w$ and $w^\prime$.
Therefore, $m_T$ in the number of misclustered graphs $\vert \mathcal{M} \vert$ is $m/2$. 
The ideal distance matrix $D$ will be of size $m\times m$ with $0$ and $\Vert w - w^\prime \Vert_{L_2}$ as entries depending on whether the samples are generated from the same graphon or not. 
For such a block matrix $D$, the two non zero eigenvalues are $\pm \dfrac{m}{2}\Vert w - w^\prime \Vert_{L_2}$. 
Therefore, $\gamma$ is $\dfrac{m}{2}\Vert w - w^\prime \Vert_{L_2}$. 
Corollary \ref{theo:spect_k2} can be derived by substituting the derived $\gamma$ in the number of misclustered graphs $\vert \mathcal{M} \vert$ in Theorem~\ref{theo:spect_error} as shown below. 

\begin{align}
    \vert \mathcal{M} \vert = \mathcal{O} \left( \dfrac{m}{ \Vert w - w^\prime \Vert_{L_2}^2 n_0^2} \right) \nonumber
\end{align}
Let us assume $\Vert w - w^\prime \Vert_{L_2} \geq C\dfrac{m}{n_0}$ where $C$ is a large constant, then as $n \rightarrow \infty$, $\dfrac{m^2 n_0^2 \log n}{n} \rightarrow 0$, $\vert \mathcal{M} \vert \rightarrow 0$.$\hfill \square$

\subsection{Similarity Based Semi-Definite Programming (SSDP)}
\label{sec:proof_ssdp}
We make Assumptions \ref{assum:lipschitz}--\ref{assum:equivalence} on the $K$ graphons 
to study the recovery of clusters from Algorithm \ref{alg:sdp_clust}.
The proof structure for cluster recovery stated in Theorem~\ref{theo:sdp_sim} is as follows:
\begin{enumerate}
    \item We establish deviation bound between the estimated similarity matrix $\widehat{S}$ and the ideal similarity matrix $S$ (Lemma \ref{lemma:S_hat_S}).
    \item We derive the recoverability condition by adapting Proposition 1 of \citet{perrot2020nearoptimal} and the obtained deviation bound (Lemma \ref{lemma:sdp_sim}).
\end{enumerate}

The ideal similarity matrix $S \in \mathbb{R}^{m\times m}$ is symmetric with $K\times K$ block structure, and $S = Z \Sigma Z^T$ where $Z \in \{0,1\}^{m\times K}$ be the clustering membership matrix and $\Sigma \in \mathbb{R}^{K\times K}$ such that $\Sigma_{ll^\prime}$ represents ideal pairwise similarity between graphs from clusters $\mathcal{C}_l$ and $\mathcal{C}_{l^\prime}$. 
From the definition of $S_{ij}$, $\Sigma_{ll^\prime} = \exp \left(-\dfrac{\Vert w_l - w_l \Vert_{L_2}}{ \sigma_l \sigma_{l^\prime}} \right)$ where $w_l$ and $w_{l^\prime}$ are graphons corresponding to clusters $\mathcal{C}_l$ and $\mathcal{C}_{l^\prime}$, respectively.
$\widehat{S}$ is the estimated similarity matrix of $S$ as mentioned earlier.
Since $\widehat{X} \in \mathbb{R}^{K\times K}$ is the normalised clustering matrix, $\widehat{X} = ZN^{-1}Z^T$ where $N$ is a diagonal matrix with $\dfrac{1}{\vert \mathcal{C}_1 \vert}, \ldots ,\dfrac{1}{\vert \mathcal{C}_K \vert}$.
We derive the deviation bound for the similarity matrices using Lemma \ref{lemma:A_W} and the result is as follows.
\begin{lemma}[Similarity deviation bound]
\label{lemma:S_hat_S}
As $n \rightarrow \infty$, $\dfrac{n_0^2\log n}{n} \rightarrow 0$, we establish
\begin{align}
    \vert \widehat{S}_{ij} - S_{ij} \vert = \mathcal{O} \left( \dfrac{1}{n_0} \right) \label{eq:S^_S}
\end{align}
 with probability converging to $1$. 
 Hence, from the result $\Vert \widehat{S} - S \Vert_F =  \mathcal{O} \left( \dfrac{m}{n_0} \right)$ with probability converging to $1$. 
\end{lemma}

\textbf{Proof of Lemma \ref{lemma:S_hat_S} (Similarity deviation bound).}
We derive the bound using Lemmas \ref{lemma:err_bound} and \ref{lemma:A_W}, and the definitions of $\widehat{S}_{ij}$ and $S_{ij}$.
\begin{align}
    \widehat{S}_{ij} &= \exp  \left( -\dfrac{\Vert A_i - A_j \Vert_F}{n_0 \sigma_i \sigma_j} \right) && \text{Consider } \sigma_i = \sigma_j = \sigma \nonumber \\
    &\stackrel{\ref{lemma:A_W}}{\geq} \exp  \left( -\dfrac{\Vert W_i - W_j \Vert_F + 4\epsilon}{n_0 \sigma^2}  \right) && \text{with probability $\rightarrow 1$ asymptotically} \nonumber \\
    &\stackrel{\ref{lemma:err_bound}}{\geq}\exp  \left( -\dfrac{\Vert w_i - w_j \Vert_{L_2} }{ \sigma^2}  \right) \exp \left( -\dfrac{ 4\epsilon + 4\sqrt{2}L}{n_0 \sigma^2} \right) && \text{$\exp(-x) \geq 1-2x$ for $x > 0$}  \nonumber \\
    &\geq S_{ij} \left( 1 -\dfrac{ 8(\epsilon+\sqrt{2}L)}{n_0 \sigma^2} \right) \nonumber \\
    &\geq S_{ij} - S_{ij} \dfrac{ 8(\epsilon+\sqrt{2}L)}{n_0 \sigma^2} && \text{$S_{ij} \in [0,1]$} \nonumber \\
    &\geq S_{ij} - \dfrac{ 8(\epsilon+\sqrt{2}L)}{n_0 \sigma^2} \label{eq:sdp_s1}
\end{align}
\begin{align}
    \widehat{S}_{ij} &= \exp  \left( -\dfrac{\Vert A_i - A_j \Vert_F}{n_0 \sigma^2} \right) \nonumber \\
    &\stackrel{\ref{lemma:A_W}}{\leq} \exp  \left( -\dfrac{\Vert W_i - W_j \Vert_F - 4\epsilon}{n_0 \sigma^2} \right) && \text{with probability $\rightarrow 1$ asymptotically} \nonumber \\
    &\stackrel{\ref{lemma:err_bound}}{\leq} \exp  \left( -\dfrac{\Vert w_i - w_j \Vert_{L_2} }{ \sigma^2}  \right) \exp \left( \dfrac{ 4\epsilon+  4\sqrt{2}L}{n_0 \sigma^2} \right) && \text{$\exp(x) \leq 1+2x$ for $x > 0$}  \nonumber \\
    &\leq S_{ij} \left( 1 +\dfrac{ 8(\epsilon+\sqrt{2}L)}{n_0 \sigma^2} \right) \nonumber \\
    &\leq S_{ij} + S_{ij} \dfrac{ 8(\epsilon+\sqrt{2}L)}{n_0 \sigma^2} && \text{$S_{ij} \in [0,1]$} \nonumber \\
    &\leq S_{ij} + \dfrac{ 8(\epsilon+\sqrt{2}L)}{n_0 \sigma^2} \label{eq:sdp_s2}
\end{align}

Thus, from \eqref{eq:sdp_s1} and \eqref{eq:sdp_s2}, we get $\vert \widehat{S}_{ij} - S_{ij} \vert \leq \dfrac{ 8(\epsilon+\sqrt{2}L)}{n_0 \sigma^2} = \mathcal{O} \left( \dfrac{1}{n_0} \right)$ for any $\epsilon$, with probability converging to $1$ as $n \rightarrow \infty$ and $\dfrac{n_0^2\log n}{ n} \rightarrow 0$.
Hence, $\Vert \widehat{S}- S \Vert_F \leq \dfrac{ 8m(\epsilon+\sqrt{2}L)}{n_0 \sigma^2} =  \mathcal{O} \left( \dfrac{m}{n_0} \right)$, with probability converging to $1$ as $n \rightarrow \infty$ and $\dfrac{m^2 n_0^2\log n}{ n} \rightarrow 0$. $\hfill \square$

The condition for exact recovery of clusters is derived by adapting Proposition 1 of \citet{perrot2020nearoptimal}. 
The proposition states the recoverability condition for such an SDP defined in \eqref{eq:sdp} in terms of the similarity deviation bound. 
Thus, we use the derived bound in Lemma \ref{lemma:S_hat_S} and establish condition on the $L_2$-distance to satisfy the proposition from \citet{perrot2020nearoptimal}. 
First, we state the adapted proposition.

We define $\Delta_1$ and $\Delta_2$ as,
$$\Delta_1 =  \min\limits_{l \neq l^\prime} ( 1-\Sigma_{ll^\prime}) \text{ \quad and \quad  } \Delta_2 = \max\limits_{ij} \vert \widehat{S}_{ij} - S_{ij} \vert.$$
Then,  the following should be satisfied for $\widehat{X}$ to be the unique optimal solution of the SDP in \eqref{eq:sdp}:
$$\Vert \widehat{S} - S \Vert_F \leq \min\limits_l |\mathcal{C}_l| \min \left \{ \dfrac{\Delta_1}{2} , \Delta_1 - 6 \Delta_2 \right \}.$$
The minimum cluster size $\min\limits_l |\mathcal{C}_l| $ in our case is $1$.
Consequently, the recoverability condition is derived and is as follows.

\begin{lemma}[Recoverability of clusters]
As $n \rightarrow \infty$, $\dfrac{m^2 n_0^2\log n}{n} \rightarrow 0$, the $\min\limits_{l \ne l^\prime} \Vert w_l - w_l^\prime \Vert_{L_2}$ should be $\Omega \left( \dfrac{m}{n_0} \right)$ so that $\widehat{X}$ is the unique optimal solution of the SDP \eqref{eq:sdp}.
\label{lemma:sdp_sim}
\end{lemma}

\textbf{Proof of Lemma \ref{lemma:sdp_sim} (Recoverability of clusters).}
We derive the condition to satisfy the stated proposition.

$\Delta_1 = 1-\max_{l \neq l^\prime} \Sigma_{ll^\prime}$ and $\Delta_2 = \min\limits_{ij} \vert \widehat{S}_{ij} - S_{ij} \vert $. The minimum cluster size in our case can be $1$.

The analyses of the two cases of the Proposition is as follows. \\
\textbf{Case 1.}
Let us assume $\Delta_2 \leq \dfrac{\Delta_1}{12}$, then $\min \left\{\frac{\Delta_1}{2} , \Delta_1 - 6\Delta_2 \right\}$ will be $\frac{\Delta_1}{2}$. Therefore,
\begin{align}
    \dfrac{8 m (\epsilon+\sqrt{2}L)}{n_0 \sigma^2} &\leq \dfrac{1}{2} - \dfrac{1}{2} \max_{l \neq l^\prime} \exp \left(-\dfrac{\Vert w_l - w_{l^\prime} \Vert_{L_2}}{ \sigma^2} \right) \nonumber \\
    \exp \left(-\min_{l \neq l^\prime} \dfrac{\Vert w_l - w_{l^\prime} \Vert_{L_2}}{ \sigma^2} \right) &\leq 1 - \dfrac{16 m (\epsilon+\sqrt{2}L)}{n_0 \sigma^2} \nonumber \\
    \min_{l \neq l^\prime} \dfrac{\Vert w_l - w_{l^\prime} \Vert_{L_2}}{ \sigma^2} &\geq  - \log \left( 1 - \dfrac{16 m (\epsilon+\sqrt{2}L)}{n_0 \sigma^2} \right) \nonumber \\
    \min_{l \neq l^\prime} \Vert w_l - w_{l^\prime} \Vert_{L_2} &\geq \sigma^2 \sum_{k=1}^\infty  \left( \dfrac{16 m (\epsilon+\sqrt{2}L)}{n_0 \sigma^2} \right)^k \dfrac{1}{k} \nonumber \\
    \min_{l \neq l^\prime} \Vert w_l - w_{l^\prime} \Vert_{L_2} &= \Omega \left( \dfrac{m}{n_0}  \right) \label{eq:case1_l2}
\end{align}

\textbf{Case 2.}
Let us assume $\Delta_2 > \dfrac{\Delta_1}{12}$, then $\min \left\{\frac{\Delta_1}{2} , \Delta_1 - 6\Delta_2 \right\}$ will be $\Delta_1 - 6\Delta_2$. Therefore,
\begin{align}
    \dfrac{8 m (\epsilon+\sqrt{2}L)}{n_0 \sigma^2} &\leq 1 - \max_{l \neq l^\prime} \exp \left(-\dfrac{\Vert w_l - w_{l^\prime} \Vert_{L_2}}{ \sigma^2} \right) - \dfrac{6*8 (\epsilon+\sqrt{2}L)}{n_0 \sigma^2} \,\, \text{with probability $\rightarrow 1$ aymptotically} \nonumber \\
    \min_{l \neq l^\prime} \dfrac{\Vert w_l - w_{l^\prime} \Vert_{L_2}}{ \sigma^2} &\geq  - \log \left( 1 - \dfrac{8 (m+6)(\epsilon+\sqrt{2}L)}{n_0 \sigma^2} \right) \nonumber \\
    \min_{l \neq l^\prime} \Vert w_l - w_{l^\prime} \Vert_{L_2} &\geq \sigma^2 \sum_{k=1}^\infty  \left( \dfrac{8 (m+6) (\epsilon+\sqrt{2}L)}{n_0 \sigma^2} \right)^k \dfrac{1}{k} \nonumber \\
    \min_{l \neq l^\prime} \Vert w_l - w_{l^\prime} \Vert_{L_2} &= \Omega \left( \dfrac{m}{n_0}  \right) \label{eq:case2_l2}
\end{align}
Thus, from \eqref{eq:case1_l2} and \eqref{eq:case2_l2}, we must satisfy $\min_{l \neq l^\prime} \Vert w_l - w_{l^\prime} \Vert_{L_2} = \Omega \left( \dfrac{m}{n_0}  \right)$ for the Proposition to hold. Consequently, Theorem \ref{theo:sdp} is the direct reflection of this lemma. $\hfill \square$

\subsection{Graph Two-Sample Testing}
Theorem \ref{theo:hyp_test} of two-sample testing is proved by deriving the probability of Type-1 and Type-2 errors. We make Assumptions \ref{assum:lipschitz}--\ref{assum:equivalence} for this case.
Let $W_1$ and $W_2$ be the $n_0\times n_0$ discretized graphons of $w_1$ and $w_2$, respectively, obtained using \eqref{eq:graphon_apprx}. Then, the alternate hypothesis $H_a$ can be rewritten using Lemma \ref{lemma:err_bound} in the following way,
\begin{align}
    \dfrac{1}{n_0}\left\Vert W_1 - W_2 \right\Vert_F + \dfrac{4\sqrt{2} L}{n_0} &\stackrel{(\ref{eq:appr_err})}{\geq} \phi \nonumber \\
    \left\Vert W_1 - W_2 \right\Vert_F &\geq n_0 \phi - 4\sqrt{2} L = \rho \label{eq:H_a}
\end{align}

We derive the probability of the errors using Lemma \ref{lemma:A_W} and is stated in the following lemmas.
\begin{lemma} [Probability of Type-1 error]
The probability of Type-1 error, i.e. rejecting the null hypothesis when it is actually true, is
\begin{align}
    \mathbb{P}(T=1 \vert H_0:True) \leq \dfrac{C}{\xi^2}\dfrac{ \log n}{n} \label{eq:type1_err}
\end{align}
where $C$ depends only on the Lipschitz constants.
\label{lemma:type1_err}
\end{lemma}

\textbf{Proof of Lemma \ref{lemma:type1_err} (Probability of Type-1 error).}
The Type-1 error is rejecting $H_0$ when it is true. 
Therefore, in this scenario, $\left\Vert W_1 - W_2 \right\Vert_F = 0$.
Thus, from Lemma \ref{lemma:A_W}, \eqref{eq:pA'_A} and \eqref{eq:pA_W}, we have $\left\Vert A_1 - A_2 \right\Vert_F \leq 4 \epsilon$ with $1 - \dfrac{C}{\epsilon^2} \dfrac{n_0^2\log n}{n} $ probability. Therefore, the probability of Type-1 error is,
\begin{align}
    \mathbb{P}(T=1 \vert H_0:True) &= \mathbb{P}(d(G_1, G_2) \geq  \xi) \nonumber \\
    &= \mathbb{P}(\left\Vert A_1 - A_2 \right\Vert_F \geq n_0 \xi) 
    && \text{err only when $n_0 \xi \leq 4\epsilon$} \nonumber \\
    &\stackrel{\eqref{eq:pA'_A}}{\leq} \dfrac{C}{\xi^2 n_0^2} \dfrac{n_0^2\log n}{n} \square \nonumber
\end{align}

\begin{lemma}[Probability of Type-2 error]
The probability of Type-2 error, i.e. accepting the null hypothesis when the alternate hypothesis is actually true, is 
\begin{align}
    \mathbb{P}(T=0 \vert H_a:True) \leq \dfrac{C_1}{\left( \phi - \frac{4\sqrt{2}C_2}{n_0}  \right)^2}\dfrac{\log n}{n} \label{eq:type2_err}
\end{align}
where $C_1$ and $C_2$ depend only on the Lipschitz constants.
\label{lemma:type2_err}
\end{lemma}

\textbf{Proof of Lemma \ref{lemma:type2_err} (Probability of Type-2 error).}
The Type-2 error is evaluating to null hypothesis when the alternate hypothesis is true. Therefore, from \eqref{eq:H_a} $\left\Vert W_1 - W_2 \right\Vert_F \geq \rho$. From Lemma \ref{lemma:A_W}, \eqref{eq:pA'_A} and \eqref{eq:pA_W},
\begin{align}
    \left\Vert A_1 - A_2 \right\Vert_F &\geq \Vert  W_1 - W_2 \Vert_F - 4\epsilon && \text{with probability $1 - \dfrac{C_1}{\epsilon^2 } \dfrac{n_0^2\log n}{n}$} \nonumber \\
    &\geq \rho - 4\epsilon \nonumber
\end{align}
The probability of Type-2 error is,
\begin{align}
     \mathbb{P}(T=0 \vert H_a:True) &= \mathbb{P}(d(G_1, G_2) <  \xi) \nonumber \\
     &= \mathbb{P}(\left\Vert A_1 - A_2 \right\Vert_F < n_0 \xi) && \text{err only when $n_0 \xi \leq \rho - 4\epsilon$; let $n_0 \xi = 4\epsilon$} \nonumber \\
     &\leq \mathbb{P}( \left\Vert A_1 - A_2 \right\Vert_F < \dfrac{\rho}{2}) \nonumber \\
    &\leq \dfrac{C_1}{\rho^2}\dfrac{n_0^2 \log n}{n} \nonumber
\end{align}
We get the probability by substituting $\dfrac{\rho}{n_0} = \phi - \dfrac{4\sqrt{2}L}{n_0}$ from \eqref{eq:H_a} in the above equation. Theorem \ref{theo:hyp_test} can be proved by asymptotic analysis of Lemmas \ref{lemma:type1_err} and \ref{lemma:type2_err}.  $\hfill \square$

\section{DSC and SSDP Algorithms}
The proposed algorithms DSC and SSDP are described as follows:

\begin{minipage}{0.48\textwidth}
\begin{algorithm}[H]
 \SetKwInOut{Input}{input} \SetKwInOut{Output}{output}
 \Input{Adjacency matrices $G_1,...,G_m$, \\ histogram size $n_0$ 
 }
 \Output{$K$ clusters $\mathcal{C}_1,...,\mathcal{C}_K$}
 \BlankLine
 \textbf{Construct distance matrix} Compute $\widehat{D} \in \mathbb{R}^{m\times m}$, where $\widehat{D}_{ij} = d(G_i,G_j)$ \\ \, \\ \, \\ 
 \textbf{Clustering} Apply spectral clustering to $\widehat{D}$ with $K$ number of clusters resulting in $\mathcal{C}_1,...,\mathcal{C}_K$\\
 \caption{Distance based Spectral Clustering (DSC)}
 \label{alg:spectral_clust}
\end{algorithm}
\end{minipage}
\hfill
\begin{minipage}{0.5\textwidth}
\begin{algorithm}[H]
 \SetKwInOut{Input}{input} \SetKwInOut{Output}{output}
 \Input{Adjacency matrices $G_1,...,G_m$, \\ histogram size $n_0$ 
 }
 \Output{$K$ clusters $\mathcal{C}_1,...,\mathcal{C}_K$}
 \BlankLine
 \textbf{Construct similarity matrix} Compute $\widehat{S} \in \mathbb{R}^{m\times m}$, where $\widehat{S}_{ij} = \exp \left(-\frac{d(G_i,G_j)}{\sigma_i \sigma_j} \right)$ 
 with $\sigma_1 = \ldots = \sigma_n$ \\
 \vspace{1mm}
 \textbf{Clustering} Find $\widehat{X}$ using \eqref{eq:sdp} and apply standard spectral clustering to $\widehat{X}$ resulting in $\mathcal{C}_1,...,\mathcal{C}_K$.
 \caption{Similarity based Spectral Clustering (SSDP)}
 \label{alg:sdp_clust}
\end{algorithm}
\end{minipage}

\section{Experimental Details}
\label{app:exp}
In this section, we present experimental details and additional experiments.

\subsection{Simulated Data - Heatmap of Graphons}
\begin{figure}[h]
\centering
\includegraphics[width=0.8\linewidth]{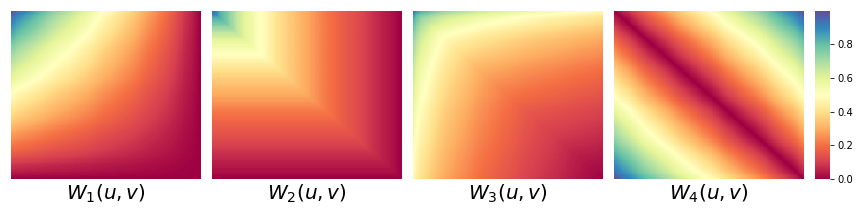}
\caption{Heatmaps of graphons $W_1$, $W_2$, $W_3$ and $W_4$.}
\label{fig:heatmap}
\end{figure}

Figure~\ref{fig:heatmap} shows the heatmap of the considered four graphons $W_1, W_2, W_3$ and $W_4$. We sample graphs from these graphons for the experiments.

\subsection{Choice of $n_0$}
\label{sec:n0}
\begin{figure}[h]
\centering
\includegraphics[width=\linewidth]{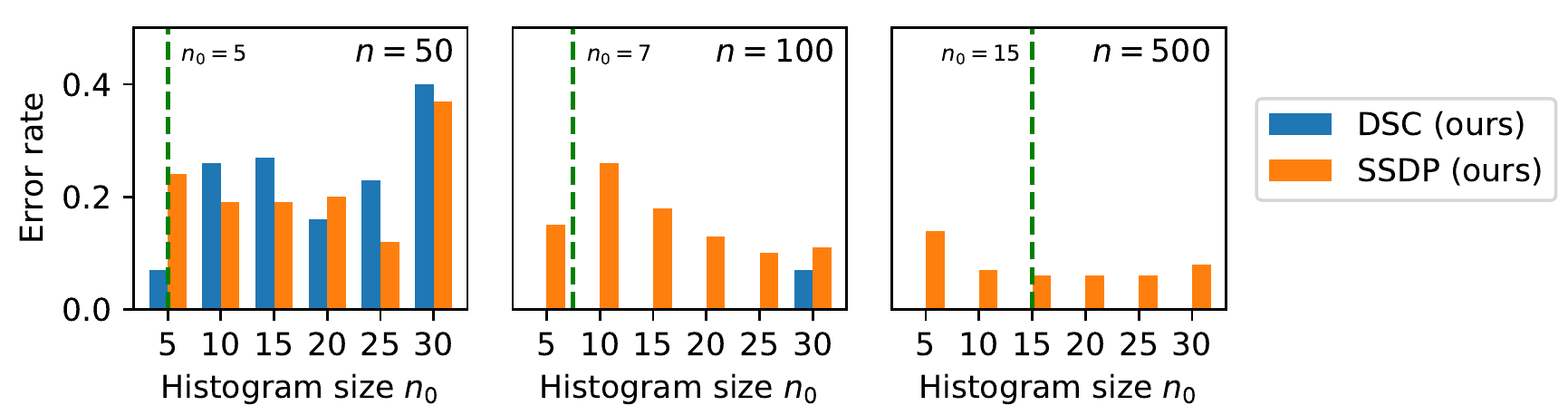}
\caption{Validation of the bound for $n_0$. The plot shows the average error rate (percentage of misclustered graphs) of the proposed algorithms DSC and SSDP for different $n=\{50,100,500\}$.}
\label{fig:n0}
\end{figure}

We validate the theoretically deduced bound for $n_0 = \mathcal{O}(\sqrt{n/\log n})$ by sampling $5$ graphs with a fixed number of nodes $n$ from each of the four graphons, in total $20$ graphs, and measuring the performance of DSC and SSDP for different $n_0= \{5,10,15,20,25,30\}$.
We perform three simulations with $n=\{50,100,500\}$ and fix neighbourhood of one in SSDP.
Figure~\ref{fig:n0} shows the average error of both the algorithms over $5$ independent trials.
Based on the theoretical considerations for $n_0$ ($\ll \sqrt{n/\log n}$), we evaluate $n_0=\{5,7,15\}$ for $n=\{50,100,500\}$, respectively.
The experimental results show that the derived bound for $n_0$ serves as a reasonable choice (if not the best) for both DSC and SSDP irrespective of $n$.
Hence, the choice of $n_0$ can be deterministic and adaptive with respect to $n$, thus making our algorithms parameter-free.

\subsection{Experimental results using Adjusted Rand Index (ARI)}
\label{sec:exp_ari}
\begin{figure}[h]
\centering
\includegraphics[scale=0.8]{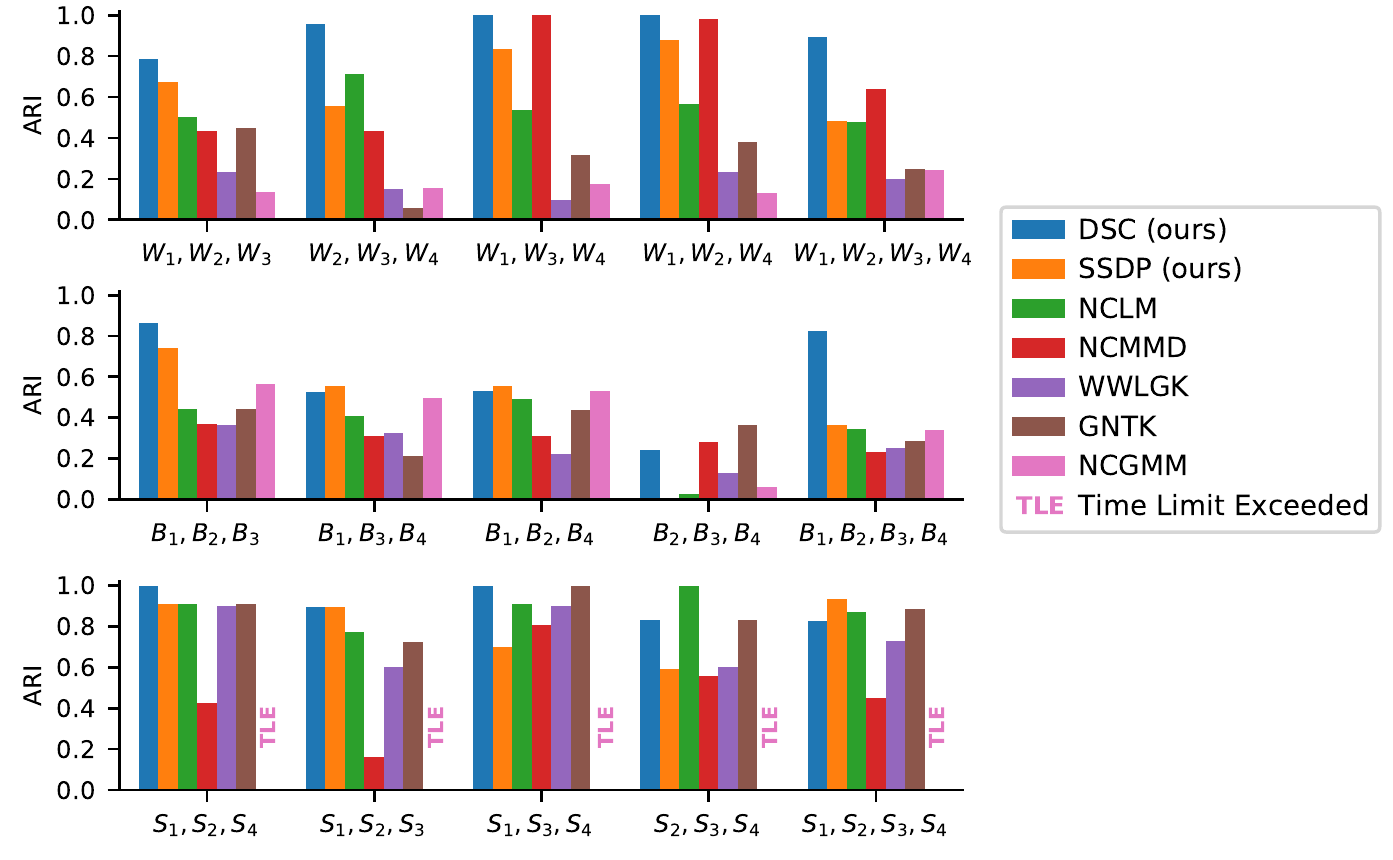}
\caption{Evaluation of DSC and SSDP with other methods. \textbf{(row 1)} Results on simulated data using ARI.
\textbf{(rows 2 and 3)} Results on real data from Bioinformatics and Social Networks, respectively. 
}
\label{fig:sim_data_ri}
\end{figure}

In this section, we provide the results for evaluation of algorithms on simulated and real data under the same setting as described in Section~\ref{sec:experiment}.
Figure~\ref{fig:sim_data_ri} shows the evaluation of all the discussed algorithms using ARI, where the observations made from error rate hold.

\begin{figure}[b]
\centering
\includegraphics[width=\linewidth]{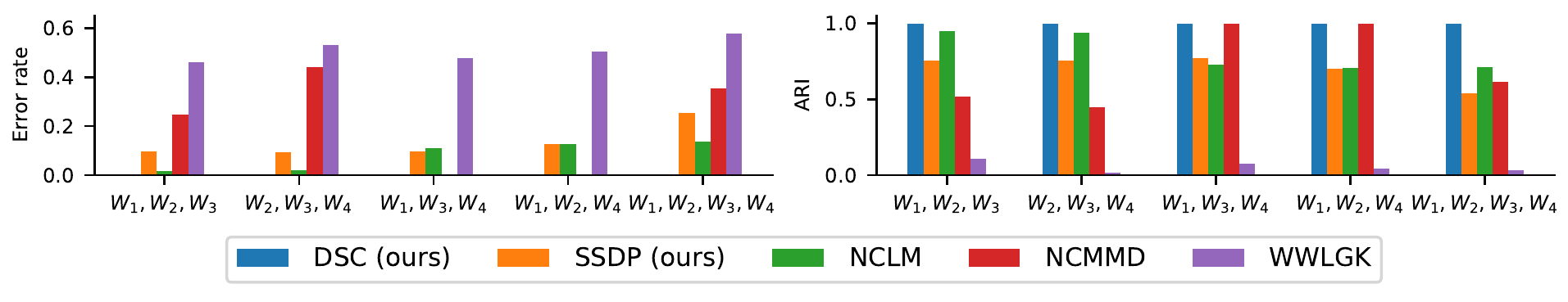}
\caption{Evaluation of all algorithms except GNTK and NCGMM using average error rate and average ARI for large simulated data.}
\label{fig:sim_data_large}
\end{figure}

\subsection{Evaluation on large simulated data}
\label{sec:large_graph_exp}
As mentioned in Section~\ref{sec:experiment}, we evaluate algorithms except GNTK and NCGMM on large graphs sampled from the four graphons $W_1,W_2,W_3$ and $W_4$ with nodes between $100$ and $1000$. 
Figure~\ref{fig:sim_data_large} shows the results measured using average error rate and average ARI, respectively.
The proposed algorithm DSC outperforms the others in all the case and SSDP stands second or third best, as observed in simulated data with small graphs.

\subsection{Computation Time of Algorithms}
\label{sec:time_all}
Table 1 shows the time (measured in seconds) taken for the algorithms on the considered dataset combinations.
Three clusters also show similar behavior to four clusters.

\begin{table}[h]
\label{tab:time}
\caption{Time measured in seconds for algorithms on different datasets combinations}
\begin{center}
\begin{tabular}{lrrrrrrr}
\multicolumn{1}{c}{\bf Dataset} & \multicolumn{1}{c}{\bf DSC} & \multicolumn{1}{c}{\bf SSDP} & \multicolumn{1}{c}{\bf NCLM} & \multicolumn{1}{c}{\bf NCMMD} & \multicolumn{1}{c}{\bf WWLGK} & \multicolumn{1}{c}{\bf GNTK} & \multicolumn{1}{c}{\bf NCGMP}
\\ \hline \\
$W_1,W_2,W_3$ & 0.13 & 0.22 & 0.17 & 11.66 & 0.66 & 225.46 & 15.43 \\
$W_2,W_3,W_4$ & 0.14 & 0.23 & 0.18 & 12.98 & 0.70 & 199.98 & 16.27 \\
$W_1,W_3,W_4$ & 0.13 & 0.25 & 0.19 & 12.93 & 0.71 & 200.85 & 16.98 \\
$W_1,W_2,W_4$ & 0.13 & 0.24 & 0.18 & 12.55 & 0.68 & 198.38 & 16.54 \\
$W_1,W_2,W_3,W_4$ & 0.20 & 0.38 & 0.28 & 22.05 & 1.19 & 390.07 & 30.18 \\
$B_1,B_2,B_3$ & 1.10 & 1.25 & 0.17 & 20.72 & 2.54 & 28.21 & 219.58 \\
$B_1,B_3,B_4$ & 0.99 & 1.18 & 0.16 & 21.51 & 2.85 & 32.22 & 225.02 \\
$B_1,B_2,B_4$ & 1.01 & 1.08 & 0.14 & 15.61 & 1.92 & 19.49 & 174.25 \\
$B_2,B_3,B_4$ & 1.04 & 1.13 & 0.13 & 11.99 & 0.78 & 13.38 & 21.47 \\
$B_1,B_2,B_3,B_4$ & 1.33 & 1.46 & 0.183 & 28.66 & 3.19 & 40.18 & 278.80 \\
$S_1,S_2,S_4$ & 1.50 & 1.65 & 8.21 & 1125.71 & 454.19 & 1609.78 & TLE \\
$S_1,S_2,S_3$ & 1.25 & 1.34 & 0.36 & 77.67 & 15.32 & 294.53 & TLE \\
$S_1,S_3,S_4$ & 1.52 & 1.64 & 8.07 & 1001.52 & 348.25 & 1485.90 & TLE \\
$S_2,S_3,S_4$ & 1.48 & 1.69 & 8.88 & 1035.98 & 440.87 & 1757.06 & TLE \\
$S_1,S_2,S_3,S_4$ & 1.97 & 2.22 & 9.47 & 1069.28 & 437.21 & 2060.68 & TLE
\end{tabular}
\end{center}
\end{table}

\subsection{Scalability experiment}
\label{sec:scalability_exp}

\begin{figure}[h]
    \centering
    \includegraphics[scale=0.8]{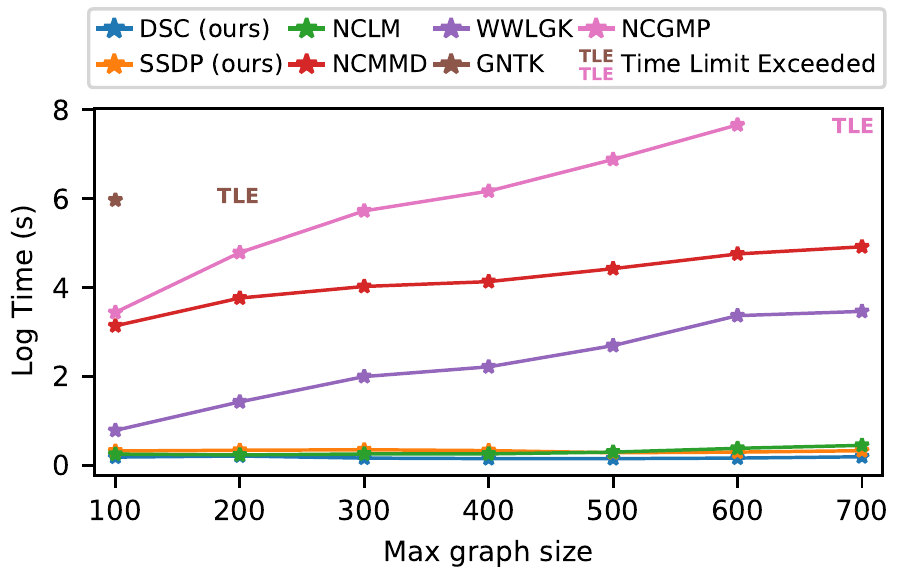}
    \caption{Computation time of algorithms on different sets of simulated data for four clusters case demonstrating the scalability of each algorithm. Computation time is plotted in log scale.}
    \label{fig:time_scalability}
\end{figure}

We evaluate the scalability of the considered algorithms using simulated data by measuring the time taken for clustering $40$ random graphs, $10$ sampled from each of the graphons $W_1,W_2,W_3$ and $W_4$.
We did $7$ experiments in which the size of the sampled graphs are varied as $[50,\text{max\_size}]$ where $\text{max\_size}=\{100,200,300,400,500,600,700\}$.
Figure~\ref{fig:time_scalability} shows the experimental results which illustrates high scalability of DSC, SSDP and NCLM over other algorithms. Note that the experiment shows NCLM as scalable as DSC and SSDP since the sampled graphs are small.

\subsection{Two-Sample Testing}
\begin{figure}[t]
    \centering
    \includegraphics[scale=0.9]{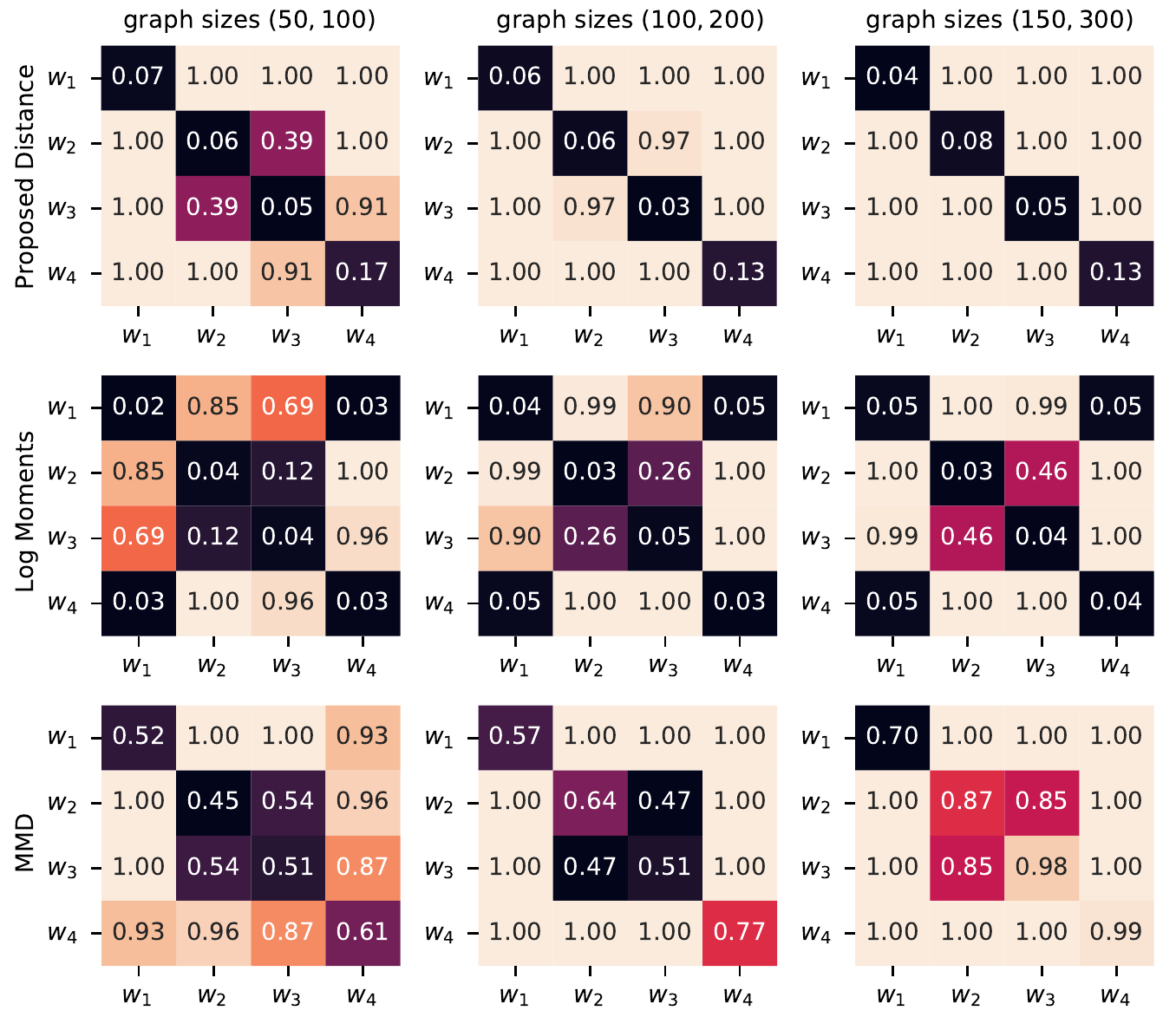}
    \caption{Illustration of two-sample testing with the proposed distance vs log moments and MMD on varying $n$ in graph pairs of size $(n,2n)$. The plots show the average power of the test $T$. Test based on the proposed distance is consistent for sufficiently large graphs and efficient compared to other methods in distinguishing even closer graphons.}
    \label{fig:testing_all}
\end{figure}

In this section, we evaluate the efficacy of the proposed test $T$ with different $d(G_1,G_2)$ by varying the graph sizes $n$. 
We consider $n=\{50,100,150\}$ and fix $n_0=10$ from the theoretical bound for evaluating the test $T$.
The power is computed using the test $T$ for the significance level $0.05$, and the plots in Figure~\ref{fig:testing_all} show the average power computed over $500$ trials of bootstrapping $100$ samples generated from all pairs of graphons for $d(G_1,G_2)$ as our proposed distance, log moments and MMD, respectively.
From the result for graph sizes $(50,100)$, we observe that the graphon pair $(W_2,W_3)$ is not easily distinguishable (low $H_0$ rejection probability), which can be explained by their respective $L_2$ distance that is shown in the left plot of Figure~\ref{fig:testing}.
This issue does not arise in testing larger graphs as the result shows for graph sizes $(100,200)$ and $(150,300)$.
Therefore, test $T$ with the proposed distance can distinguish between pairs of graphons that are quite close provided that the observed graphs are sufficiently large, thus proving to be consistent. 
On the other hand, log moments and MMD based tests show weakness in distinguishing the graphons, where log moments based test $T$ accepts the null hypothesis in most cases even when the graphons are different for all graph sizes. For instance, the result for graphon pair $W_1$ and $W_4$ is indistinguishable using log moments statistic for any graph size.
On the contrary, MMD based test $T$ rejects the null hypothesis almost always for larger graphs (diagonal values in all graph cases). 
Thus, we conclude that the proposed test $T$ in \ref{eq:test} is consistent and this experiment illustrates the efficiency of the test $T$ compared to other plausible test statistics.

\begin{figure}[t]
    \centering
    \includegraphics[scale=0.9]{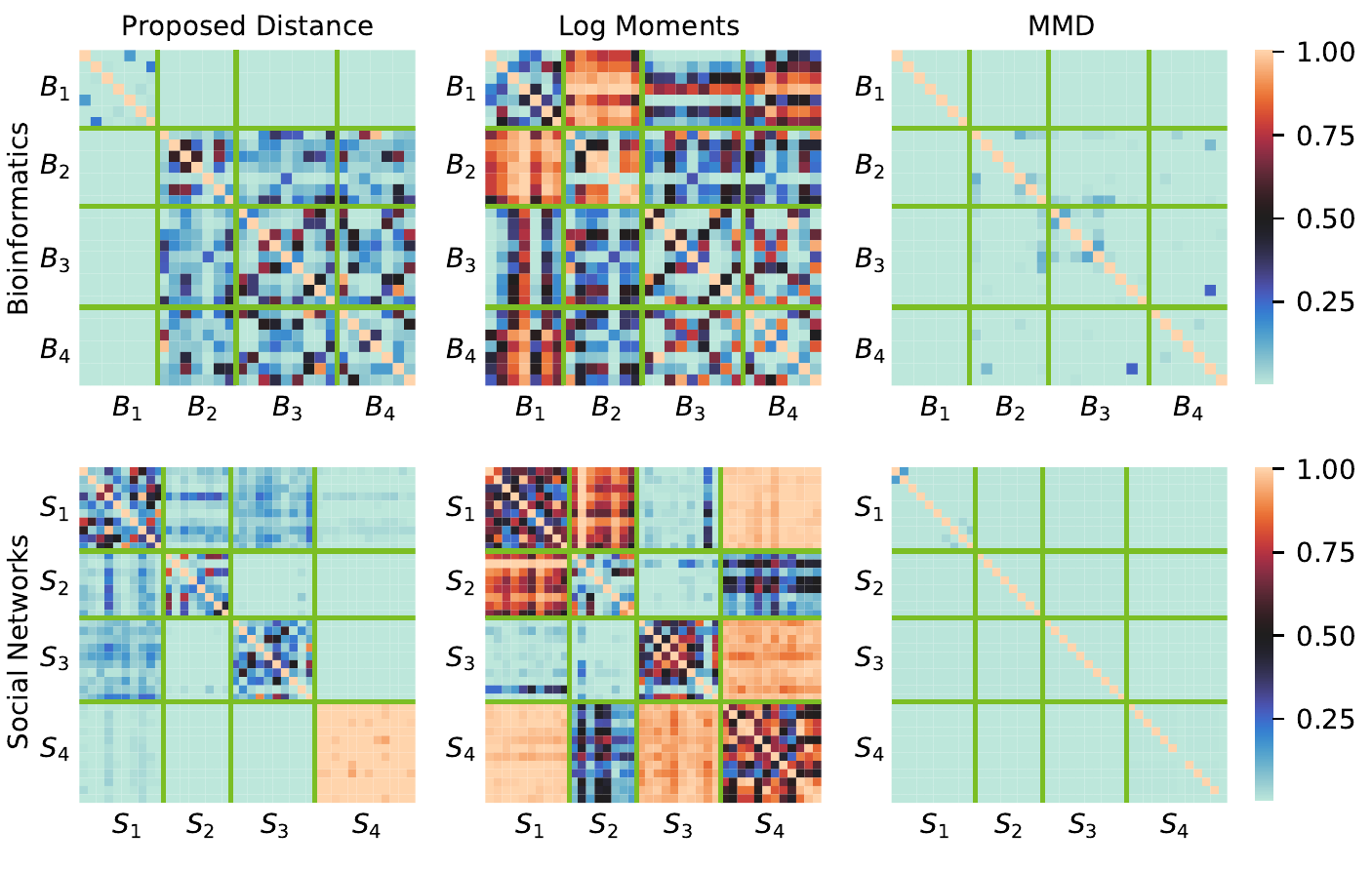}
    \caption{Illustration of two-sample testing with the proposed distance vs log moments and MMD on real datasets - Bioinformatics and Social Networks. The plots show the $p$-value of the test $T$. Test based on the proposed distance is better than the other two tests.}
    \label{fig:testing_all_real}
\end{figure}

Subsequently, we evaluate the efficacy of the above tests on the discussed real datasets -- Bioinformatics and Social Networks.
We consider graphs from a dataset to belong to a population and hence the objective of the test statistic is to distinguish graphs from different populations, that is, graphs from two different datasets.
Since the populations are not known and the real graphs are treated as representatives of the population, we compute $p$-value of the test instead of power to measure the efficacy.
The $p$-value is the evidence for rejecting the null hypothesis which implies that the smaller the $p$-value, the stronger the evidence that the null hypothesis should be rejected.
Therefore,  the $p$-value should be high (greater than the significance level) for graphs from same population and low ($\simeq 0$) for graphs from different populations.
Figure~\ref{fig:testing_all_real} shows the result for both the dataset cases and different tests.
From the results, it is clear that the log moments and MMD based tests are poor and inefficient on real datasets as log moments based test has high acceptance of null hypothesis for almost all the pair of graphs from any population and MMD based test rejects the null hypothesis always except when the graphs are the same.
Whereas, the test using our proposed distance perform well on large graphs from Social Networks datasets, for instance, $S_4$ with other datasets and within itself. 
This test performs well even for small graphs when compared to the other two tests.

\end{document}